\documentclass[sn-mathphys]{sn-jnl}

\jyear{2022}%

\theoremstyle{thmstyleone}%
\theoremstyle{thmstyletwo}%

\theoremstyle{thmstylethree}%

\usepackage{amsmath}
\usepackage{amssymb}
\usepackage{amsfonts}
\usepackage{latexsym}
\usepackage{lipsum}
\usepackage{multicol}
\usepackage{accents}
\usepackage{wrapfig}
\usepackage{float}
\usepackage{tabularx}
\usepackage{euscript}
\usepackage{color}
\usepackage{xcolor}
\usepackage{graphicx}
\usepackage{float}
\usepackage{subcaption}
\usepackage{epsfig}
\usepackage{hyperref}
\usepackage{setspace}
\usepackage{fancyhdr}
\usepackage[T1]{fontenc} 
\usepackage{hyperref}
\usepackage[utf8x]{inputenc} 
\graphicspath{{Figures/}}
\begin{document}

\title[Automatic Hyperparameter Optimization Using Genetic Algorithm in Deep Reinforcement Learning for Robotic Manipulation Tasks]{Automatic Hyperparameter Optimization Using Genetic Algorithm in Deep Reinforcement Learning for Robotic Manipulation Tasks}


\author[1,2]{\fnm{Adarsh} \sur{Sehgal}}\email{asehgal@nevada.unr.edu}

\author[1,2]{\fnm{Nicholas} \sur{Ward}}\email{nhward@nevada.unr.edu}

\author*[1,2]{\fnm{Hung} \sur{La}}\email{hla@unr.edu}

\author[2]{\fnm{Sushil} \sur{Louis}}\email{sushil@cse.unr.edu}

\affil[1]{\orgdiv{
Advanced Robotics and Automation (ARA) Laboratory}}

\affil[2]{\orgdiv{
Department of Computer Science and Engineering}, \orgname{University of Nevada}, \orgaddress{\city{Reno}, \postcode{89557}, \state{NV}, \country{USA}}}


\abstract{Learning agents can make use of Reinforcement Learning (RL) to decide their actions by using a reward function. However, the learning process is greatly influenced by the elect of values of the hyperparameters used in the learning algorithm. This work proposed a Deep Deterministic Policy Gradient (DDPG) and Hindsight Experience Replay (HER) based method, which makes use of the Genetic Algorithm (GA) to fine-tune the hyperparameters' values. This method (GA+DDPG+HER) experimented on six robotic manipulation tasks: \textit{FetchReach}; \textit{FetchSlide}; \textit{FetchPush}; \textit{FetchPick\&Place}; \textit{DoorOpening}; and \textit{AuboReach}. Analysis of these results demonstrated a significant increase in performance and a decrease in learning time. Also, we compare and provide evidence that GA+DDPG+HER is better than the existing methods. }




\maketitle

\section{Introduction}\label{sec1}
Reinforcement Learning (RL) \cite{sutton1998introduction} has recently been applied to a variety of applications, including robotic table tennis \cite{9560764}, surgical robot planning \cite{9635867}, rapid motion planning in bimanual suture needle regrasping \cite{9561673}, and Aquatic Navigation \cite{9635925}. Each one of these applications employs RL as a motivating alternative to automating manual labor.

Studies have shown that hyperparameter tuning, particularly when utilizing machine learning, can have a significant impact on an algorithm's performance \cite{probst2019tunability, zhang2021importance}. From this came the inspiration to enhance the RL algorithm. While the hill climbers \cite{sebag1997society} can be useful in specific situations, they are not very useful in complex situations like robotic tasks employing RL. This is due to the lack of a clear correlation between performance changes and changing the settings of hyperparameters. The number of epochs it takes the learning agent to learn a given robotic
task can be used to assess its performance. Due to the non-linear nature of the relationship between the two, we looked into using a Genetic Algorithm (GA) \cite{davis1991handbook}, which may be utilized for hyperparameter optimization. Even while GAs can have significant computational costs, which could be an overhead for any RL algorithm, GAs offer a potential method to tune the hyperparameters in comparable kinds of issues once and for all. GA has the ability to adjust the hyperparameters once and utilize them endlessly in subsequent RL trainings, saving computational resources that would otherwise be spent each time an RL is applied to an agent. \cite{nguyen2014review} demonstrates how GA is preferred over other optimization techniques. This is due to GA's capacity to handle either discrete or continuous variables, or both. It enables the simultaneous evaluation of n members of a population, enabling multiprocessor machines to run parallel simulations. GA is inherently suitable for the solution of multi-objective optimization problems since it works with a population of solutions \cite{deb1999multi}. \cite{wetter2004comparison} evaluated a small number of cost functions to compare the performance of eight algorithms in solving simple and complex building models. They discovered that the GA frequently approached the optimal minimum. This demonstrates that GA is a powerful algorithm for solving complex problems, making it our top pick for hyperparameter optimization.

\cite{sehgal2019deep}, \cite{sehgal2019genetic}, \cite{sehgal2022ga}, and \cite{10.1007/978-3-030-33723-0_29} are some of the closely related works. These publications' findings add to the growing body of data that using a GA to automatically modify the hyperparameters for DDPG+HER can greatly enhance efficiency. The discrepancy can have a notable impact on how long it takes a learning agent to learn. 

We develop a novel automatic hyperparameter tweaking approach in this paper, which we apply to DDPG+HER from \cite{baselines}. The algorithm is then applied to four existing robotic manipulator gym environments as well as three custom-built robotic manipulator gym environments. Furthermore, the entire algorithm is examined at various stages to determine whether the technique is effective in increasing the overall efficiency of the learning process. The final results support our claim and provide sufficient evidence that automating the hyperparameter tuning procedure is critical since it reduces learning time by up to $57\%$. Finally, we compare GA+DDPG+HER to four different approaches used on \textit{FetchReach}. GA+DDPG+HER outperforms them all. Open source code is available at \textcolor{orange}{\href{https://github.com/aralab-unr/ga-drl-aubo-ara-lab}{https://github.com/aralab-unr/ga-drl-aubo-ara-lab}}.

The following is a list of our major contributions: 
\begin{itemize}
  \item A novel algorithm for automatic hyperparameter tweaking.
  \item Six simulated and one real task were used to test the algorithm. To analyze the algorithm, we created \textit{Aubo-i5} simulated and actual custom environments.
  \item As the GA develops, the training process is evaluated based on a variety of hyperparameters.
  \item The efficiency of DDPG+HER was tested over ten runs using GA+DDPG+HER discovered hyperparameters in both simulated and actual manipulation tasks.
  \item GA+DDPG+HER was compared to other approaches.
\end{itemize}

\section{Background}\label{sec2}

\subsection{Genetic Algorithm (GA)} 
Genetic Algorithms (GAs) \cite{davis1991handbook, holland1992genetic, goldberg1988genetic} were created to explore poorly-understood areas \cite{de1988learning}, where an exhaustive search is impossible and other search methods perform poorly. GAs, when employed as function optimizers, aim to maximize fitness that is linked to the optimization goal. On a range of tough design and optimization issues, evolutionary computing techniques in general, and GAs in particular, have had a lot of empirical success. They begin with a population of candidate solutions that have been randomly initialized and are commonly encoded in a string (chromosome). A selection operator narrows the search space to the most promising locations, whereas crossover and mutation operators provide new potential solutions.

To choose parents for crossover and mutation, we employed ranking selection \cite{goldberg1991comparative}. Higher-ranked (fitter) individuals are probabilistically selected through rank selection. Unlike fitness proportionate selection, ranking selection is concerned with the existence of a fitness difference rather than its magnitude. Uniform crossover \cite{syswerda1989uniform} is used to create children, who are then altered by flip mutation \cite{goldberg1988genetic}. Binary coding with concatenated hyperparameters is used to encode chromosomes. \cite{sehgal2019lidar} shows one such example of GA paired with Lidar-monocular visual odometry (LIMO).

\subsection{Deep Reinforcement Learning (DRL)}
Q-learning \cite{watkins1992q} approaches have been used by autonomous robots to do a range of tasks \cite{La_TCST2015}, and significant study has been done in this field since its inception \cite{watkins1992q}, with some work focused on continuous action spaces \cite{gaskett1999q, doya2000reinforcement, hasselt2007reinforcement, baird1994reinforcement} and others on discrete action spaces \cite{wei2017discrete}. Reinforcement Learning (RL) \cite{sutton1998introduction} has been used to improve locomotion \cite{kohl2004policy, endo2008learning} and manipulation \cite{peters2010relative, kalakrishnan2011learning}. In the field of robotic research, deep reinforcement learning (DRL) has become a potent control method \cite{vargas2019creativity}. The DRL is more strong in its detailed examination of the environment than in control theory. When used on robots, this DRL capacity produces more intelligent and human-like behavior. Robots can thoroughly examine the environment and discover useful solutions when DRL techniques are used in conjunction with adequate training \cite{kober2013reinforcement}.

There are two types of RL methods: off-policy and on-policy. On-policy methods seek to assess or improve the policy upon which decisions are made, for e.g. SARSA learning \cite{andrew1999reinforcement}. For real robot systems, off-policy methods \cite{munos2016safe} such as the Deep Deterministic Policy Gradient algorithm (DDPG) \cite{lillicrap2015continuous}, Proximal Policy Optimization \cite{schulman2017proximal}, Advantage Actor-Critic (A2C) \cite{mnih2016asynchronous}, Normalized Advantage Function algorithm (NAF) \cite{gu2016continuous} are useful. There is also a lot of work on robotic manipulators \cite{deisenroth2011learning, 7864333}. Some of this work relied on fuzzy wavelet networks \cite{lin2009h}, while others relied on neural networks \cite{miljkovic2013neural, duguleana2012obstacle} to complete their goals. \cite{nguyen2019review} provides a comprehensive overview of modern deep reinforcement learning (DRL) algorithms for robot handling.
Goal-conditioned reinforcement learning (RL) frames each activity in terms of the intended outcome and, in theory, has the potential to teach a variety of abilities \cite{chane2021goal}. The robustness and sample effectiveness of goal-reaching procedures are frequently enhanced by the application of hindsight experience replay (HER) \cite{andrychowicz2017hindsight}. 
For our trials, we are employing DDPG in conjunction with HER. \cite{nguyen2018deep} describes recent work on applying experience ranking to increase the learning pace of DDPG+HER.

Both a single robot \cite{pham2018, Pham_SSRR2018} and a multi-robot system \cite{La_CYBER2013,  La_SMCA2015, pham2018cooperative, Dang_MFI2016, rahimi2018} have been extensively trained/taught using RL. Both model-based and model-free learning algorithms have been studied previously. Model-based learning algorithms are heavily reliant on a model-based teacher to train deep network policies in real-world circumstances.

Similarly, there has been a lot of work on GA's \cite{davis1991handbook} \cite{deb2002fast} and the GA operators of crossover and mutation \cite{poon1995genetic}, which have been applied to a broad variety of problems. GA has been used to solve a wide range of RL problems \cite{liu2009study,moriarty1999evolutionary,mikami1994genetic,poon1995genetic}.

\subsection {Proximal Policy Optimization (PPO) }

Policy Gradient (PG) \cite{sutton1999policy} approaches are one of several RL techniques used to optimize the policy. These methods employ the rewards to compute an estimator of the policy gradient and plug it into a stochastic gradient ascent algorithm \cite{alagha2022target} . The Proximal Policy Optimization (PPO) \cite{schulman2017proximal} technique can be applied to environments with either continuous or discrete action spaces \cite{wu2019investigation}. For reinforcement learning, PPO is a family of Policy Gradient algorithms \cite{melo2019learning}. Without any prior knowledge, it is a model-free reinforcement learning. A DRL agent is said to be model-free if it hasn't learned an explicit model of its surroundings. A surrogate objective for computing policy updates is the main component of the PPO algorithm. In the spirit of a trust region strategy, the surrogate objective regulates significant policy updates so that each step stays within a near vicinity of the previous-iteration policy \cite{hsu2020revisiting}. 

\subsection {Advantage Actor Critic (A2C)}
Advantage actor critic, often known as A2C, is a synchronous variant of Asynchronous advantage actor-critic (A3C) models that performs as well as or better than the asynchronous version \cite{schulman2017proximal}. A2C \cite{mnih2016asynchronous} technique is based on policy gradients. By forecasting what actions are good in advance, it dissociates value estimates from action selection. Only effective revisions to the policy are produced using value estimation. In essence, it immediately raises the likelihood of positive behaviors and lowers the likelihood of negative ones \cite{scholte2022goal}. \cite{haydari2021impact} The policy estimation and value function estimate procedures used in actor-critic based DRL models apply to an advantage function. The A2C approach estimates the policy function with a critic network and the Q-value function with an actor network instead of just using a single learner neural network to estimate the Q-value function. Actor and critic networks are updated simultaneously using A2C actor-critic models. The calculation time is longer for A3C models since they estimate both actor and critic networks simultaneously and asynchronously. We would be using A2C in our experiments because there is not much of a performance difference between synchronous and asynchronous actor-critic models.

\subsection{GA on DDPG+HER} 
GA can be used to solve a variety of optimization problems as a function optimizer. This study concentrates on the DDPG+HER, which was briefly discussed earlier in this chapter. Based on their fitness values, GAs can be utilized to optimize the hyperparameters in the system. GA tries to achieve the highest level of fitness. Various mathematical formulas can be used to convert an objective function to a fitness function.

Existing DDPG+HER algorithms have a set of hyperparameters that cannot be changed. When GA is applied to DDPG+HER, it discovers a better set of hyperparameters, allowing the learning agent to learn more quickly. The fitness value for this problem is the inverse of the number of epochs. GA appears to be a potential technique for improving the system's efficiency.

\section[GA with DDPG+HER]{Genetic Algorithm optimization for Deep Reinforcement Learning}\label{sec3}

\subsection{Reinforcement Learning}

Consider a typical RL system, which consists of a learning agent interacting with the environment. \cite{andrychowicz2017hindsight} $S$ is the set of states, $A$ is the set of actions, $p(s_0)$ is a distribution of initial states, $r: S \times A \xrightarrow{} R$ is a reward function, $p(s_{t+1}\vert s_t,a_t)$ are transition probabilities, and $\gamma \in [0,1]$ is a discount factor that can be used to represent an environment. 

A deterministic policy depicts the relationship between states and actions as follows: $\pi : S \xrightarrow{} A$. Every episode begins with the sampling of the initial state $s_0$. The agent acts at for each timestep $t$ based on the current state st: $s_t$: $a_t = \pi(s_t)$.
The accomplished action is rewarded with $r_t = r(s_t, a_t)$, and the distribution $p(.\vert s_t,a_t)$ aids in sampling the new state of the environment. $R_t = \sum_{i=T}^\infty \gamma^{i-t}r_i$ is the discounted sum of future rewards. The purpose of the agent is to maximize its expected return $E[R_t\vert s_t,a_t]$, and an optimal policy can be defined as any policy $\pi^*$, such that $Q^{\pi^*} (s,a) \geq Q^\pi (s,a)$ for each $s \in S, a \in A$, and any policy $\pi$. An optimal Q-function, $Q^*$, is a policy that has the same Q-function as the best policy and fulfills the \textit{Bellman} equation:
  
\begin{gather} 
    Q^*(s,a) = E_{s'~p(.\vert s,a))} [r(s,a) + \gamma \underset{a'\in A}{max} Q^*(s',a'))]. \label{equation}
\end{gather}

\subsection{Deep Deterministic Policy Gradients (DDPG)}

There are two neural networks in  \textit{Deep Deterministic Policy Gradients} \cite{lillicrap2015continuous} \textit{(DDPG)}: an Actor and a Critic. The critic neural network is an action-value function approximator $Q:S\times A\xrightarrow{}R$ and the actor neural network is a target policy $\pi :S\xrightarrow{}A$. With weights $\theta^Q$ and $\theta^\mu$, the critic network $Q(s,a\vert \theta^Q)$ and actor network $\mu(s\vert \theta^\mu)$ are randomly initialized.

By offering a reliable target to the training process, the target network, which is typically created as a replica of the critic network, is also used for the purpose of updating the critic network. To produce episodes, a behavioral policy is used, which is a noisy variation of the target policy, $\pi_b(s)=\pi(s) + \mathcal{N}(0,1)$. A critic neural network is trained similarly to a DQN, except that the target $y_t$ is computed as $y_t=r_t+\gamma Q(s_{t+1},\pi(s_{t+1}))$, where $\gamma$ is the discounting factor. The actor network is trained using the loss $\mathcal{L}_a=-E_aQ(s,\pi(s))$. The experience replay and target network are both critical for sustaining the DDPG method's training and enabling deep neural networks.

\subsection{Hindsight Experience Replay (HER)}

Hindsight Experience Reply (HER) \cite{andrychowicz2017hindsight} aims to learn from setbacks by imitating human behavior. Even if the agent does not achieve the intended goal, the agent learns from each experience. HER considers the modified goal to be whatever condition the agent reaches. Only the transition $(s_t||g,a_t,r_t,s_{t+1}||g)$ with the original aim $g$ is saved in standard experience replay. The transition $(s_t||g',a_t,r'_t,s_{t+1}||g')$ to modified goal $g'$ is also stored by HER. HER performs admirably with extremely scarce incentives and is far superior to shaped rewards in this regard.

\subsection{Open Problem Discussion}

The efficiency of DDPG + HER is a concern. Using a better set of hyperparameters in the algorithm can increase the performance of the majority of robotic activities. The number of epochs for the learning agent to learn specific robotic tasks can be used to assess performance. The remaining sections of this section demonstrate how changing the values of various hyperparameters has a major impact on the agent's competency. Later in this section, the solution to this problem is described, and the accompanying experimental findings show that the proposed approach outperforms the existing reinforcement learning technique.

\subsection{DDPG + HER and GA}

The main contribution of our paper is presented in this section: the genetic algorithm explores the space of hyperparameter values utilized in DDPG + HER for values that maximize task performance while reducing the number of training epochs. The following hyperparameters are targeted: discounting factor $\gamma$; polyak-averaging coefficient $\tau$ \cite{polyak1992acceleration}; learning rate for critic network $\alpha_{critic}$; learning rate for actor-network $\alpha_{actor}$; percent of times a random action is taken $\epsilon$; and standard deviation of Gaussian noise added to not completely random actions as a percentage of the maximum absolute value of actions on different coordinates $\eta$. All of the hyperparameters have a range of 0-1, which may be justified using the equations in this section.

Adjusting the values of hyperparameters did neither boost nor reduce the agent's learning in a linear or immediately visible pattern, according to our experiments. As a result, a basic hill climber is unlikely to identify optimal hyperparameters. We use our GA to optimize these hyperparameter values because GAs were created for such poorly understood problems.

We use $\tau$, the polyak-averaging coefficient, to show the performance non-linearity for different values of $\tau$. As seen in Equation (\ref{tau}), $\tau$ is employed in the algorithm:

\begin{gather} 
    \theta^{Q'} \xleftarrow{} \tau\theta^Q+(1-\tau)\theta^{Q'},
    \nonumber\\
    \theta^{\mu'} \xleftarrow{} \tau\theta^\mu+(1-\tau)\theta^{\mu'}.
    \label{tau}
\end{gather}

Equation (\ref{target y}) demonstrates how $\gamma$ is employed in the DDPG + HER method, whereas Equation (\ref{q learning}) explains the Q-Learning update. $\alpha$ denotes the rate at which you are learning. This update equation is used to train networks.

\begin{gather} 
    y_i = r_i + \gamma Q'(s_{i+1},\mu'(s_{t+1}\vert\theta^{\mu'})\vert\theta^{Q'}), \label{target y}
\end{gather}

\begin{gather} 
    Q(s_t,a_t) \xleftarrow{} Q(s_t,a_t) + \alpha[r_{t+1} + \gamma Q(s_{t+1},a_{t+1}) \nonumber\\ - Q(s_t,a_t)].
    \label{q learning}
\end{gather}

We will need two learning rates, one for the actor-network ($\alpha_{actor}$) and the other for the critic network ($\alpha_{critic}$) because we have two types of networks. The use of the percent of times that a random action is executed $\epsilon$ is explained by equation (\ref{action}).

\begin{gather} 
a_t = 
\begin{cases}
     a^*_t \qquad \qquad \qquad \; \; \; \; \; \, with\;probability\;1 - \epsilon, \\
     random\;action \qquad  with\;probability\;\epsilon.
\end{cases}
\label{action}
\end{gather}

\begin{figure}[!h]
\centering
  \begin{subfigure}[b]{\linewidth}
    \centering
    \includegraphics[width=8cm,height=6cm]{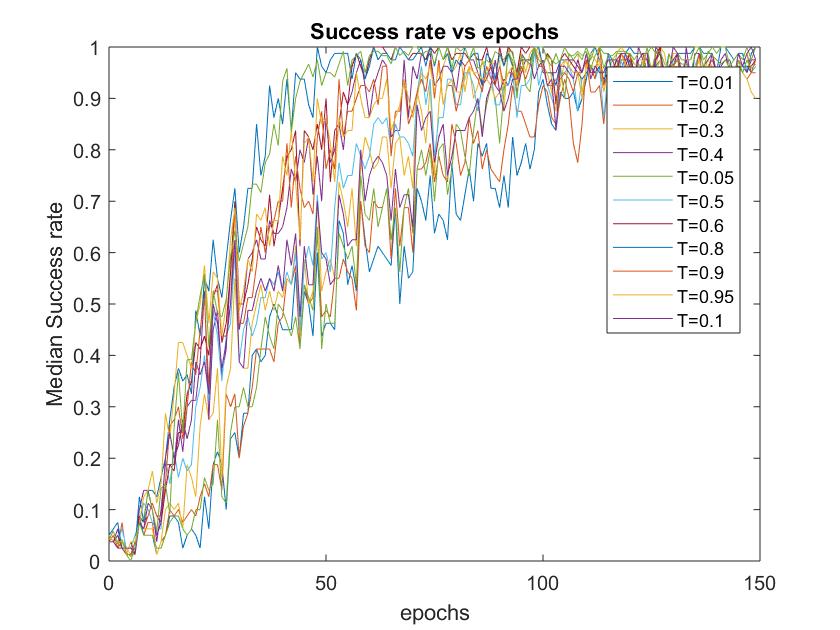}
  \end{subfigure}
  \begin{subfigure}[b]{\linewidth}
    \centering
    \includegraphics[width=8cm,height=6cm]{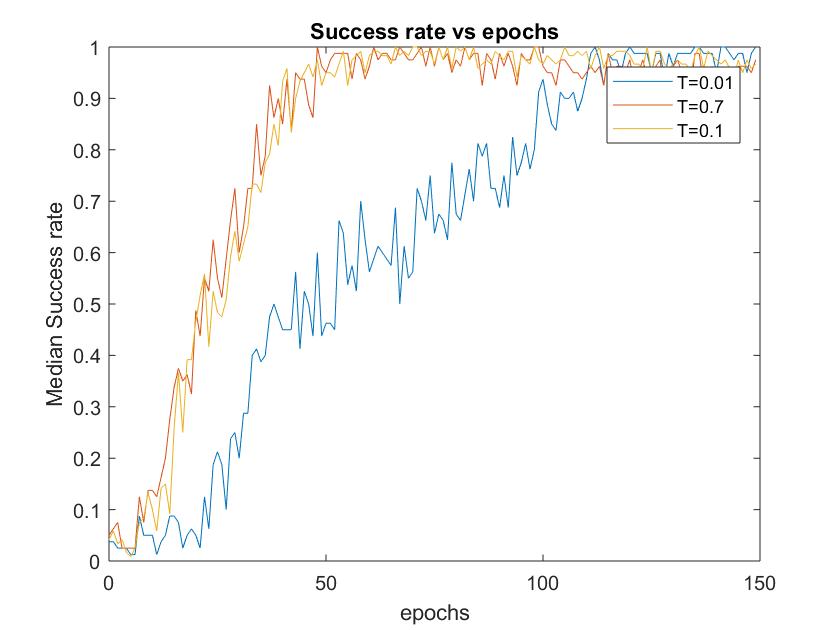}
  \end{subfigure}
  \begin{subfigure}[b]{\linewidth}
    \centering
    \includegraphics[width=8cm,height=6cm]{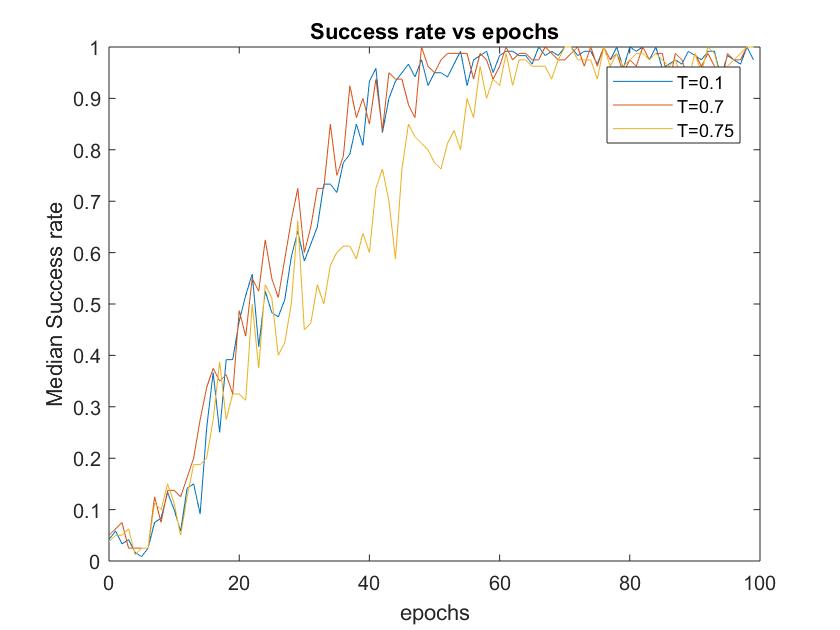}
  \end{subfigure}
  \caption{Success rate vs. epochs for various $\tau$ for \textit{FetchPick\&Place-v1} task. }
  \label{fig:TargetReaching}
\end{figure}

Figure \ref{fig:TargetReaching} indicates that changing the value of $\tau$ causes a change in the agent's learning, underscoring the importance of using a GA. The comparison of 11 $\tau$ values is shown in the first plot. For clarity, two additional charts compare three $\tau$ values. We are using four CPUs and the original (untuned) value of $\tau$ in DDPG was 0.95. All values are taken into account up to two decimal places to examine how the success rate changes as the hyperparameter values change. We can see from the plots that there is a lot of room for improvement from the original success rate.

\begin{algorithm}[!h]
\caption{Proposed GA+DDPG+HER Algorithm}\label{euclid}
\begin{algorithmic}[1]
\State Choose a population of $n$ chromosomes
\State Set the values of hyperparameters into the chromosome
\State Run the DDPG +  HER to get the number of epochs for which the algorithm first reaches success rate $\geq 0.85$
\For{all chromosome values} 
    \State Initialize DDPG
    \State Initialize replay buffer $R \gets \phi$
    \For{episode=1, M}
        \State Sample a goal $g$ and initial state $s_0$
        \For{t=0, T-1}
            \State Sample an action $a_t$ using DDPG behavioral policy
            \State Execute the action $a_t$ and observe a new state $s_{t+1}$
        \EndFor
        \For{t=0, T-1}
            \State $r_t:=r(s_t,a_t,g)$
            \State Store the transition $(s_t||g,a_t,r_t,s_{t+1}||g)$ in $R$
            \State Sample a set of additional goals for replay $G:=S$(\textbf{current episode})
            \For{$g'\in G$}
                \State $r':=r(s_t,a_t,g')$
                \State Store the transition $(s_t||g',a_t,r',s_{t+1}||g')$ in $R$ 
            \EndFor
        \EndFor
        \For{t=1, N}
            \State Sample a minibatch $B$ from the replay buffer $R$
            \State Perform one step of optimization using $A$ and minibatch $B$
        \EndFor
    \EndFor
    \State \textbf{return} $1/epochs$
\EndFor
\State Perform Uniform Crossover
\State Perform Flip Mutation at a rate of 0.1
\State Repeat for the required number of generations to find an optimal solution
\end{algorithmic}
\label{algo:ddpgherga}
\end{algorithm}

The integration of DDPG + HER with a GA, which uses a population size of 30 across 30 generations, is explained in Algorithm \ref{algo:ddpgherga}. The number of experiments with various values was used to determine population size and the number of generations. These values are sufficient to compute the success rate for a sufficient number of chromosomes. To choose parents, we used \textit{ranking selection} \cite{goldberg1991comparative}. The parents are determined probabilistically based on rank, which is determined by relative fitness (performance). Then, using \textit{uniform crossover}
\cite{syswerda1989uniform}, children are created. We are also utilizing \textit{flip mutation} \cite{goldberg1988genetic} with a 0.1 mutation frequency. Each hyperparameter is encoded using a binary chromosome, and the bits are concatenated to generate a chromosome for the GA.
Polyak-averaging coefficient, discounting factor, learning rate for critic network, learning rate for actor-network, percent of times a random action is taken, and standard deviation of Gaussian noise added to not completely random actions as a percentage of the maximum absolute value of actions on different coordinates are the six hyperparameters in order. We need 66 bits for six hyperparameters, since each hyperparameter requires 11 bits to be expressed to three decimal places. Domain-independent crossover and mutation string operators can then generate new hyperparameter values using these string chromosomes. Because tiny changes in hyperparameter values produce significant variations in success rates, we examined hyperparameter values up to three decimal places. A step size of 0.001 is, for example, considered the greatest fit for our problem. It might be important to mention that every chromosome was examined exactly once. Even if the results of identical chromosomal runs may vary slightly, we think it is safe to assume that the fitness function has a low variance, which keeps the calculation cost low.

The inverse of the number of epochs it takes for the learning agent to achieve close to maximum success rate ($\geq0.85$) for the first time determines the fitness of each chromosome (set of hyperparameter values). Because GA always maximizes the objective function, fitness is inversely proportional to the number of epochs, converting our minimization problem into a maximization problem. We utilize a GA search since an exhaustive search of the $2^{66}$-size search area is not practicable due to the length of each fitness evaluation.

\section{Experimental Results}\label{sec4}

\subsection{Experimental setup}

\begin{figure}
\centering
  \begin{subfigure}[b]{\linewidth}
   \centering
    \includegraphics[width=5.5cm,height=4cm]{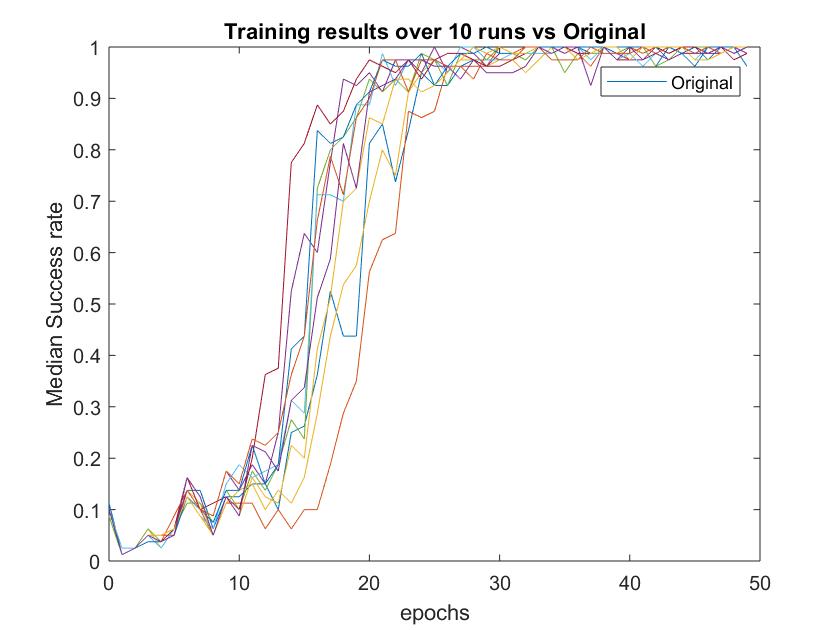}
    \caption{GA+DDPG+HER over ten runs, vs. Original}\label{fig:compareFetchPusha}
  \end{subfigure}
  \begin{subfigure}[b]{\linewidth}
  \centering
    \includegraphics[width=5.5cm,height=4cm]{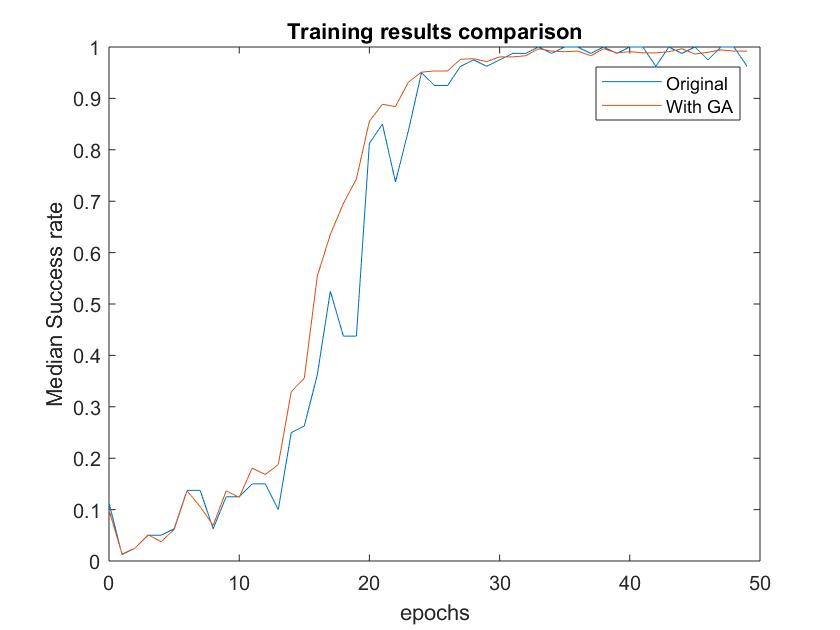}
    \caption{GA+DDPG+HER averaged over ten runs, vs. Original}\label{fig:compareFetchPushb}
  \end{subfigure}
  \caption{Success rate vs. epochs for \textit{FetchPush-v1} task when $\tau$ and $\gamma$ are found using the GA.}
  \label{fig:compareFetchPush}
\end{figure}

\begin{figure}
\centering
  \begin{subfigure}[b]{0.8\linewidth}
  \centering
    \includegraphics[width=8.6cm,height=6.5cm]{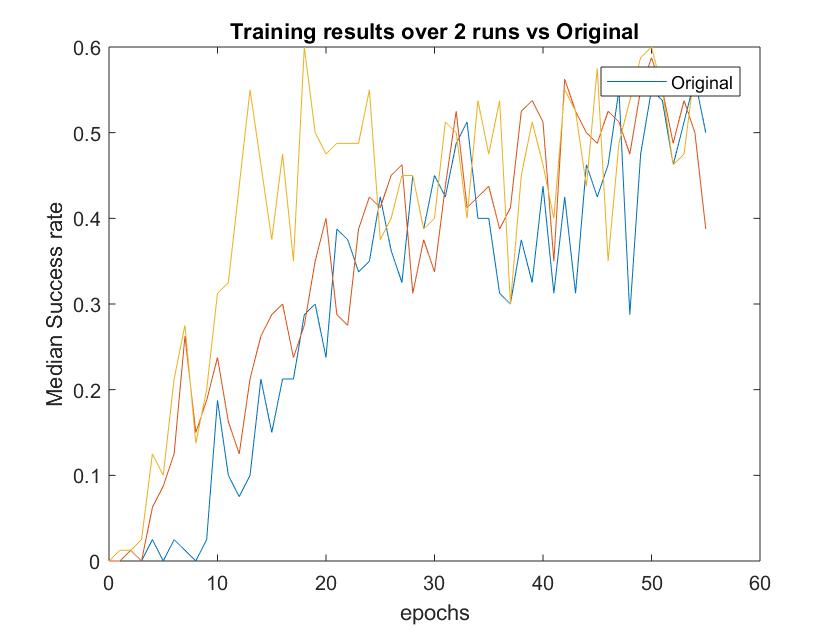}
    \caption{GA+DDPG+HER over 2 runs, vs. Original}\label{fig:plotallrunsfetchslide-1}
  \end{subfigure}
  \begin{subfigure}[b]{0.9\linewidth}
  \centering
    \includegraphics[width=8.6cm,height=6.5cm]{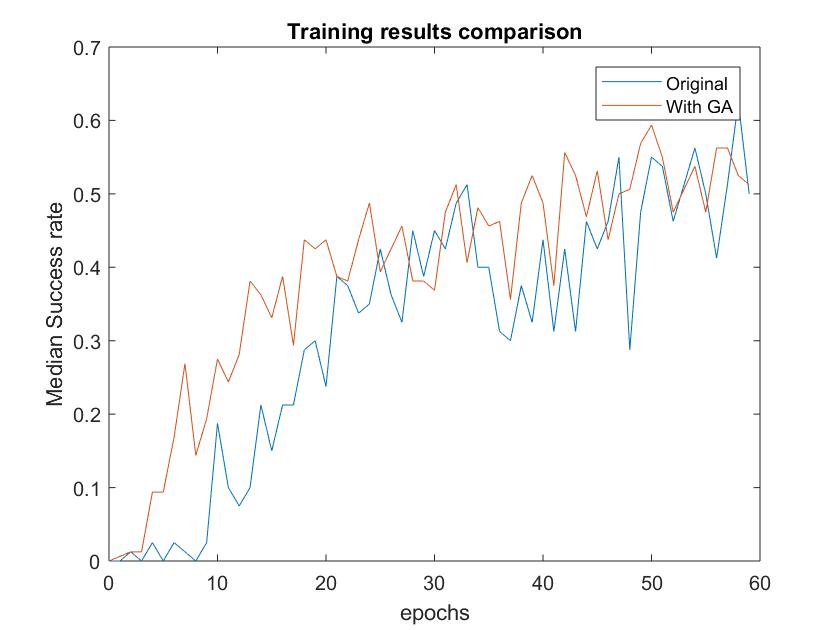}
    \caption{GA+DDPG+HER averaged over 2 runs, vs. Original}\label{fig:plotaveragefetchslide}
  \end{subfigure}
  \caption{Success rate vs. epochs for \textit{FetchSlide-v1} task when $\tau$ and $\gamma$ are found using the GA.}
  \label{fig:compareFetchSlide}
\end{figure}

\begin{figure}
\centering
  \includegraphics[width=8cm,height=2.2cm]{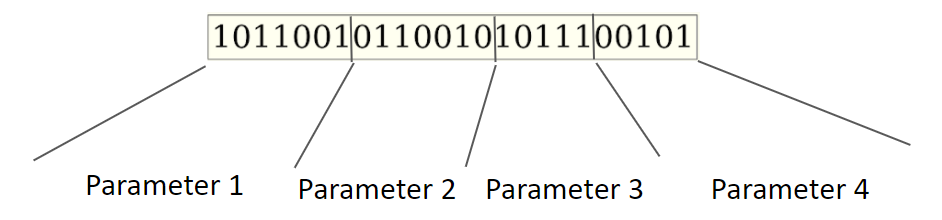}
  \caption{Chromosome representation for the GA.}
  \label{fig:chromosome_rep}
\end{figure}

A chromosome is binary encoded, as previously stated. Each chromosomal string is the result of combining all of the GA's arguments. Figure \ref{fig:chromosome_rep} depicts an example chromosome with four binary encoded hyperparameters.

\begin{figure}[!h]
\centering
\begin{multicols}{2}
    \includegraphics[width=5.5cm,height=4cm]{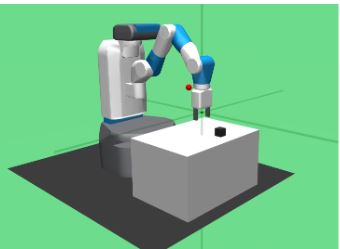}
    \subcaption{FetchPick\&Place environment}
    \vspace{0.3cm}
    \includegraphics[width=5.5cm,height=4cm]{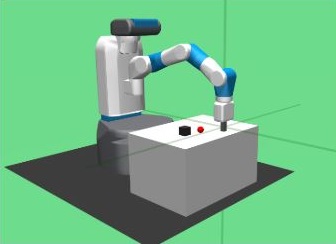}
    \subcaption{FetchPush environment}
    \vspace{0.3cm}
    \includegraphics[width=5.5cm,height=4cm]{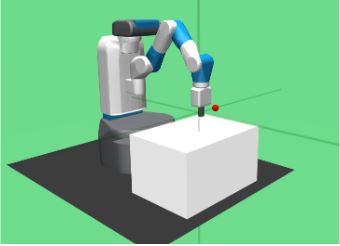}
    \subcaption{FetchReach environment}
    \includegraphics[width=5.5cm,height=4cm]{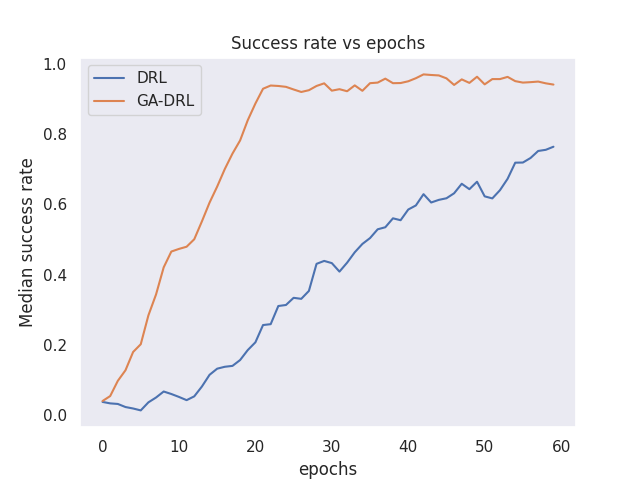}
    \subcaption{FetchPick\&Place plot}
    \includegraphics[width=5.5cm,height=4cm]{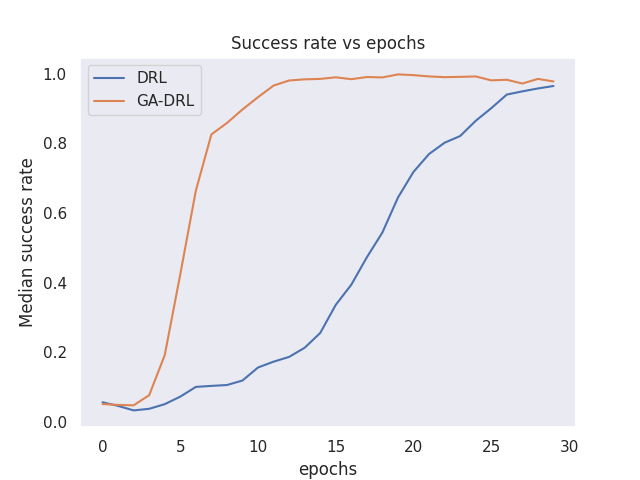}
    \subcaption{FetchPush plot}
    \includegraphics[width=5.5cm,height=4cm]{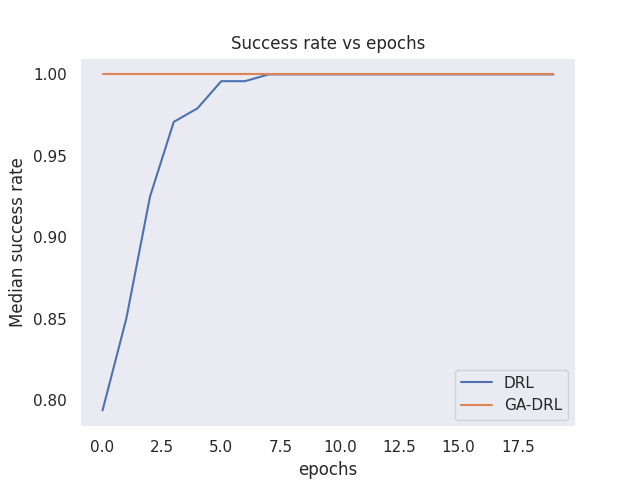}
    \subcaption{FetchReach plot}
\end{multicols}
\caption{When all six hyperparameters are identified by GA, the matching DDPG+HER versus GA+DDPG+HER charts are generated. Over ten runs, all graphs are averaged.}
\label{finalPlots}
\end{figure}

\begin{figure}
\centering
\begin{multicols}{2}
    \includegraphics[width=5.5cm,height=4cm]{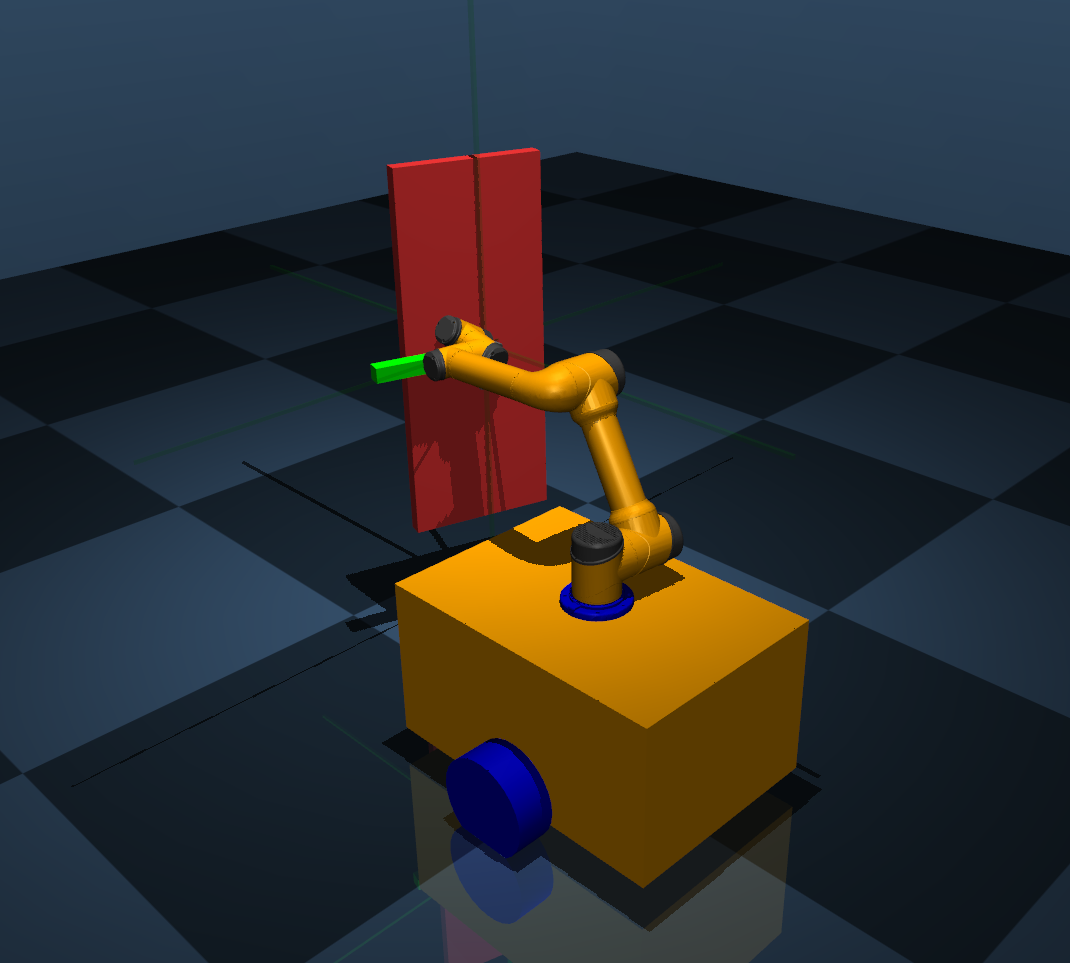}
    \subcaption{Door Opening environment}
    \vspace{0.3cm}
    \includegraphics[width=5.5cm,height=4cm]{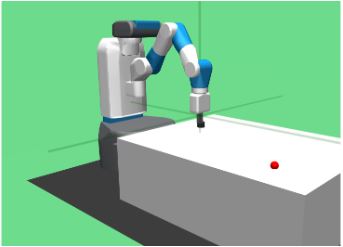}
    \subcaption{FetchSlide environment}
    \includegraphics[width=5.5cm,height=4cm]{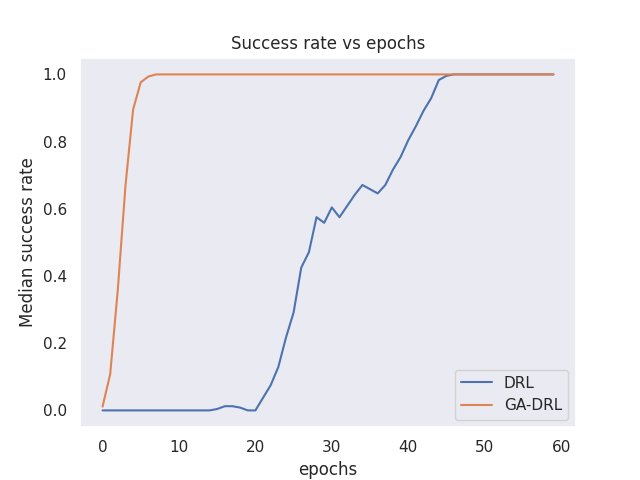}
    \subcaption{DoorOpening plot}
    \includegraphics[width=5.5cm,height=4cm]{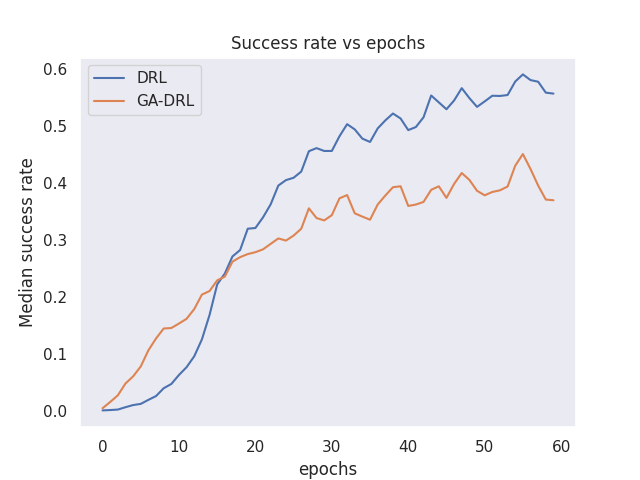}
    \subcaption{FetchSlide plot}
\end{multicols}
\caption{When all six hyperparameters are found by GA, the environments and related DDPG+HER versus GA+DDPG+HER charts are shown. Over ten runs, the graphs are averaged.}
\label{finalPlotsDoorOpening}
\end{figure}

\begin{figure}
\centering
  \begin{subfigure}[b]{0.43\linewidth}
    \centering
    \includegraphics[width=3.8cm,height=3.5cm]{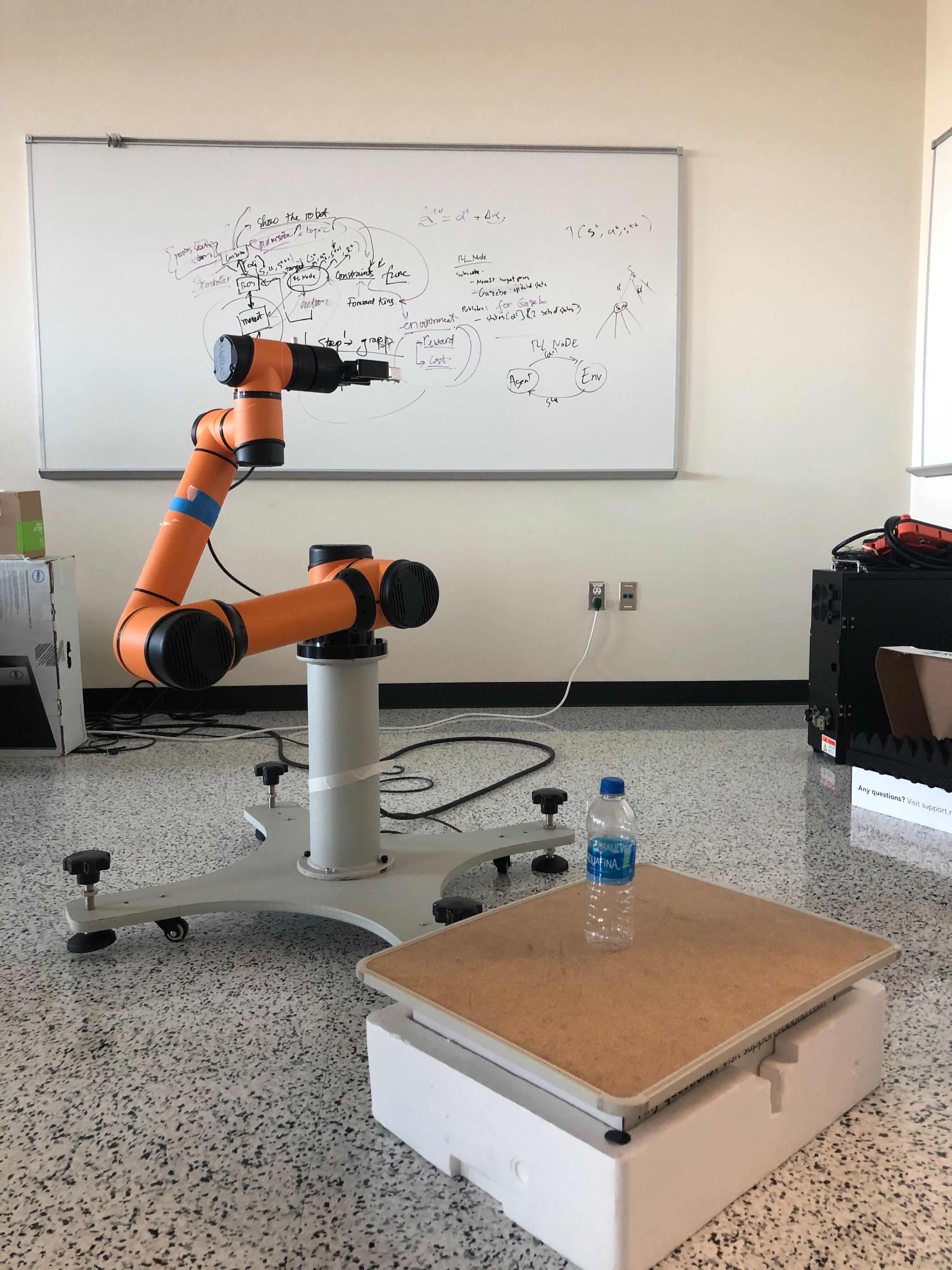}
  \end{subfigure}
  \begin{subfigure}[b]{0.43\linewidth}
    \centering
    \includegraphics[width=3.8cm,height=3.5cm]{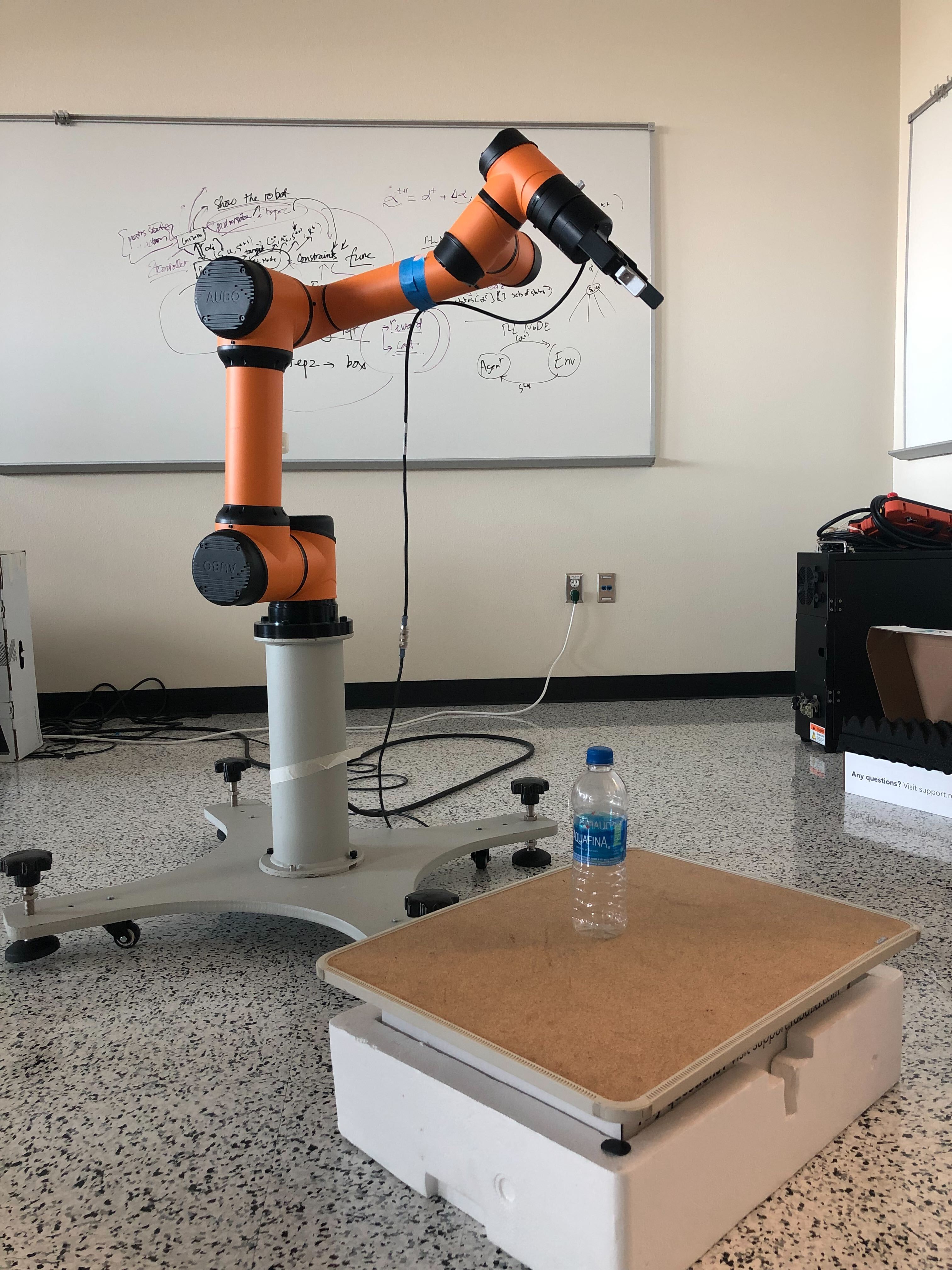}
  \end{subfigure}
  \begin{subfigure}[b]{0.43\linewidth}
    \centering
    \includegraphics[width=3.8cm,height=3.5cm]{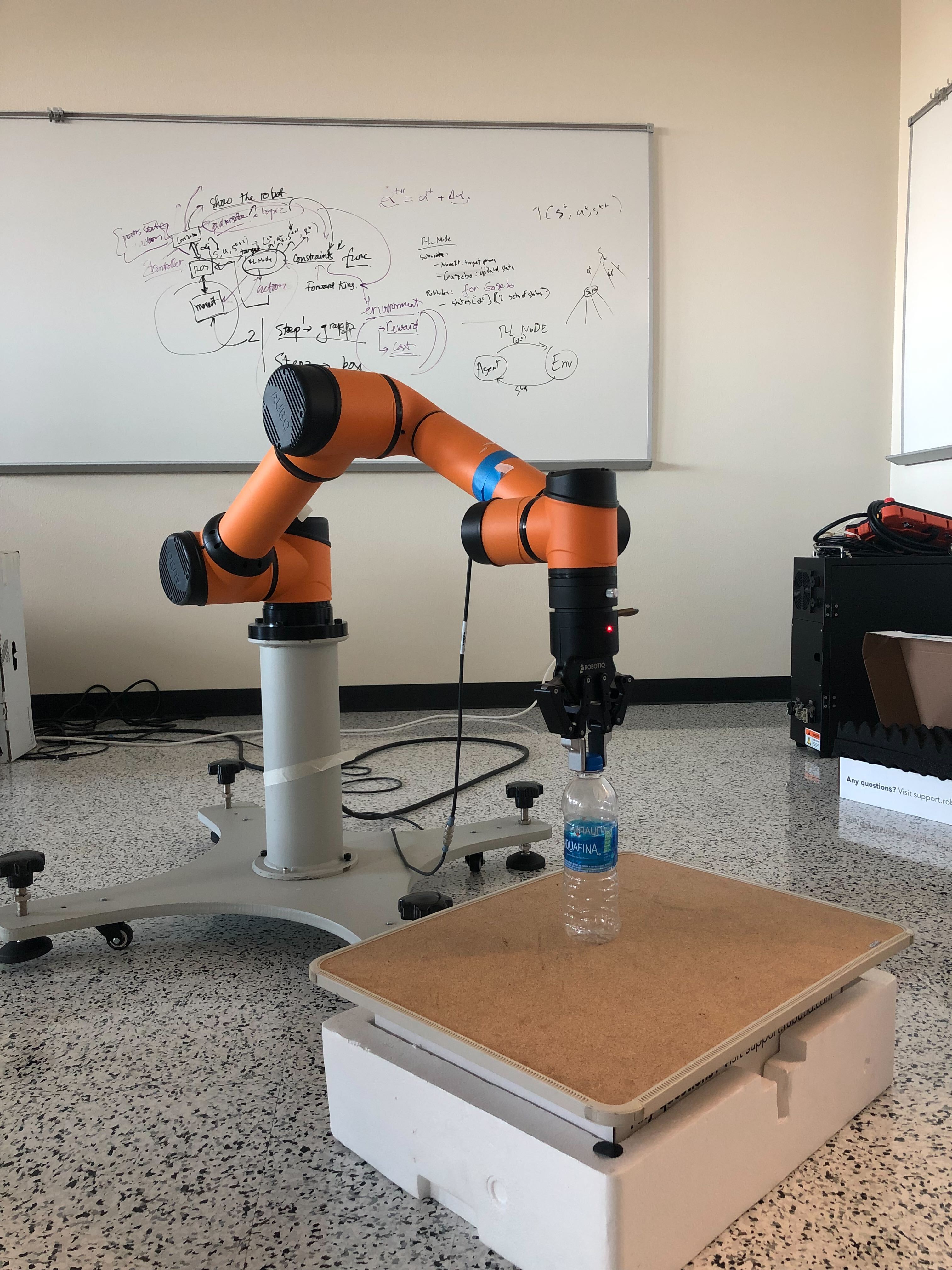}
  \end{subfigure}
  \begin{subfigure}[b]{0.43\linewidth}
    \centering
    \includegraphics[width=3.8cm,height=3.5cm]{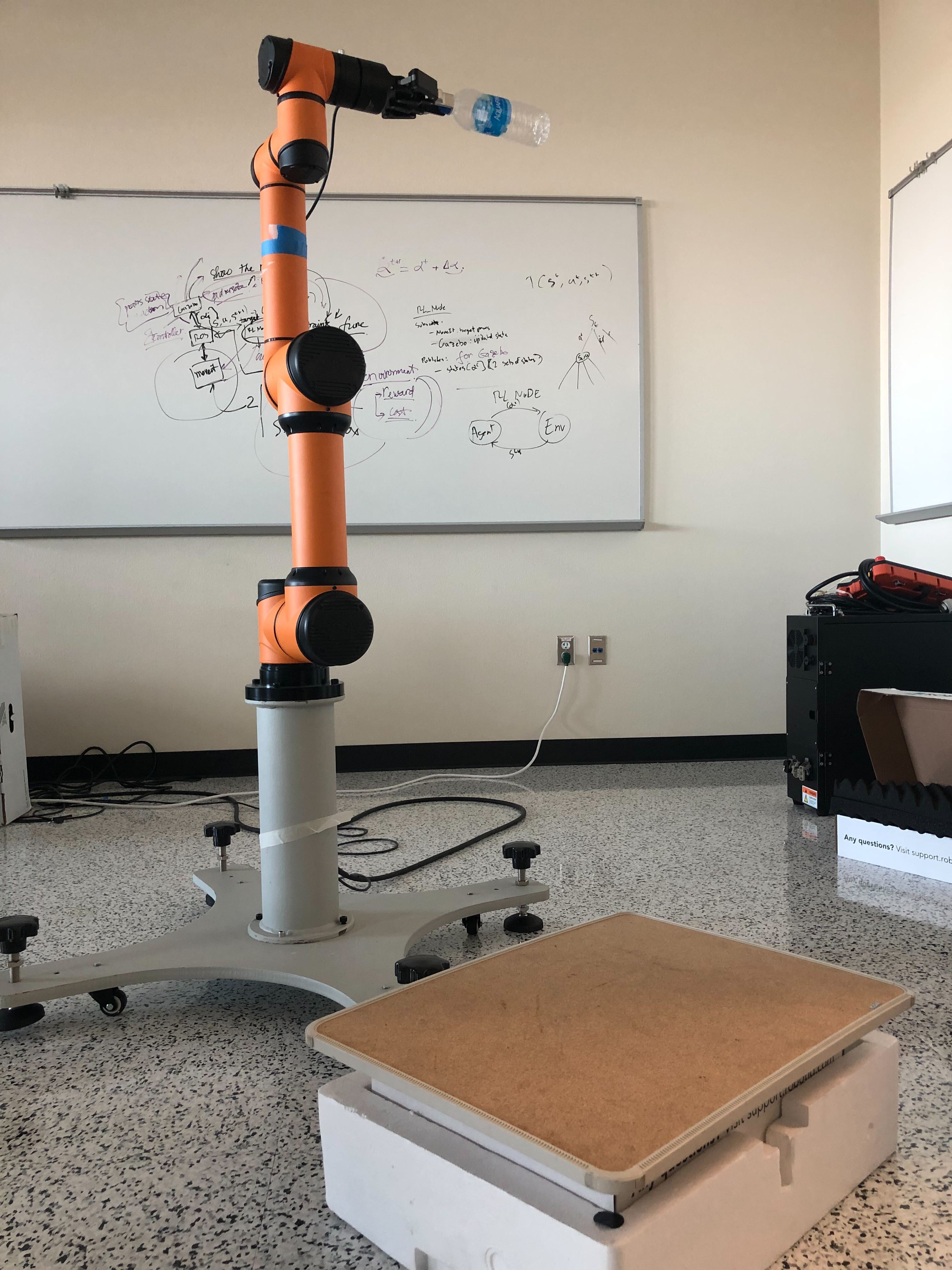}
  \end{subfigure}
  \caption{Using the most accurate policy learned via GA+DDPG+HER, the AuboReach environment performs a task in a real experiment.}
  \label{fig:auboReachEnv}
\end{figure}

\begin{figure}
\centering
  \begin{subfigure}[b]{0.43\linewidth}
    \centering
    \includegraphics[width=3.8cm,height=3cm]{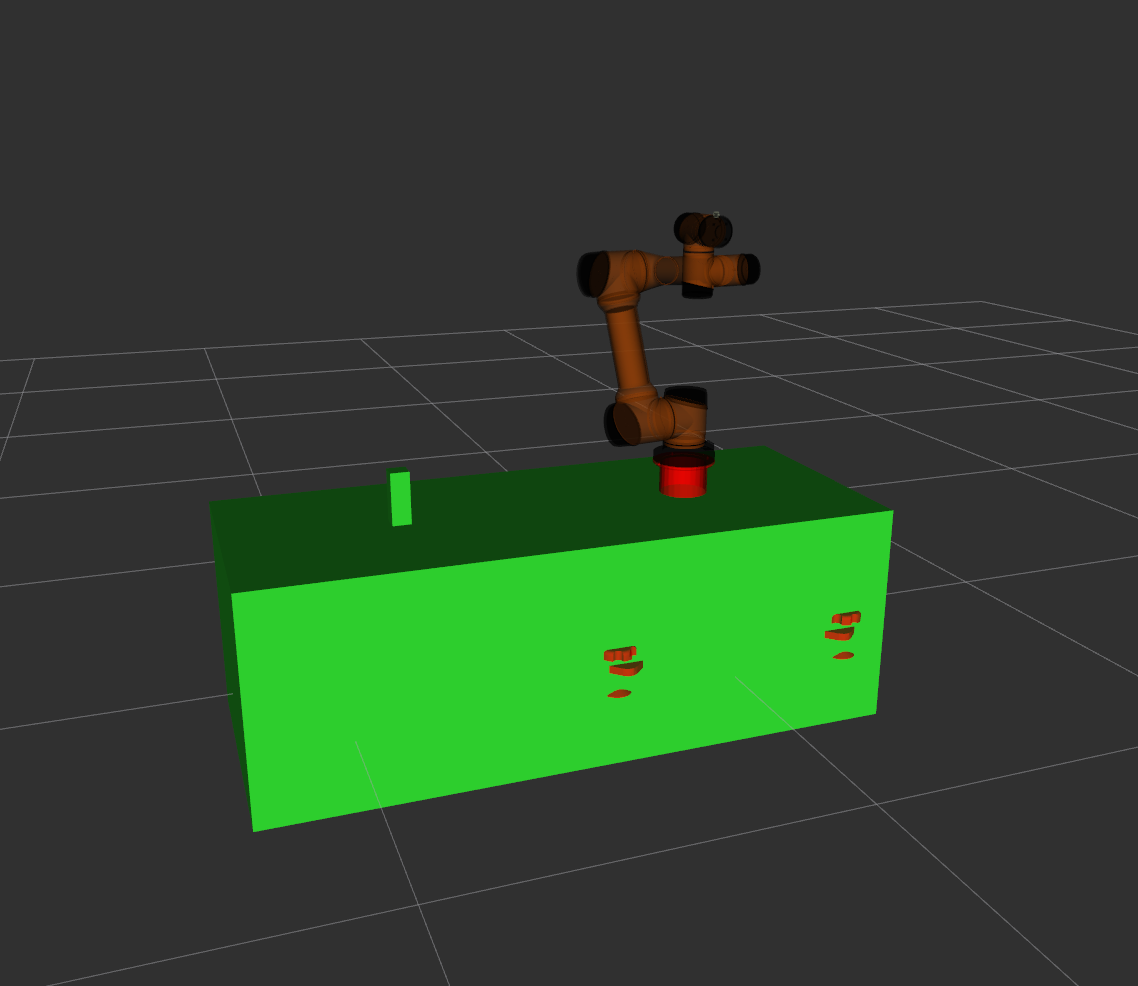}
  \end{subfigure}
  \begin{subfigure}[b]{0.43\linewidth}
    \centering
    \includegraphics[width=3.8cm,height=3cm]{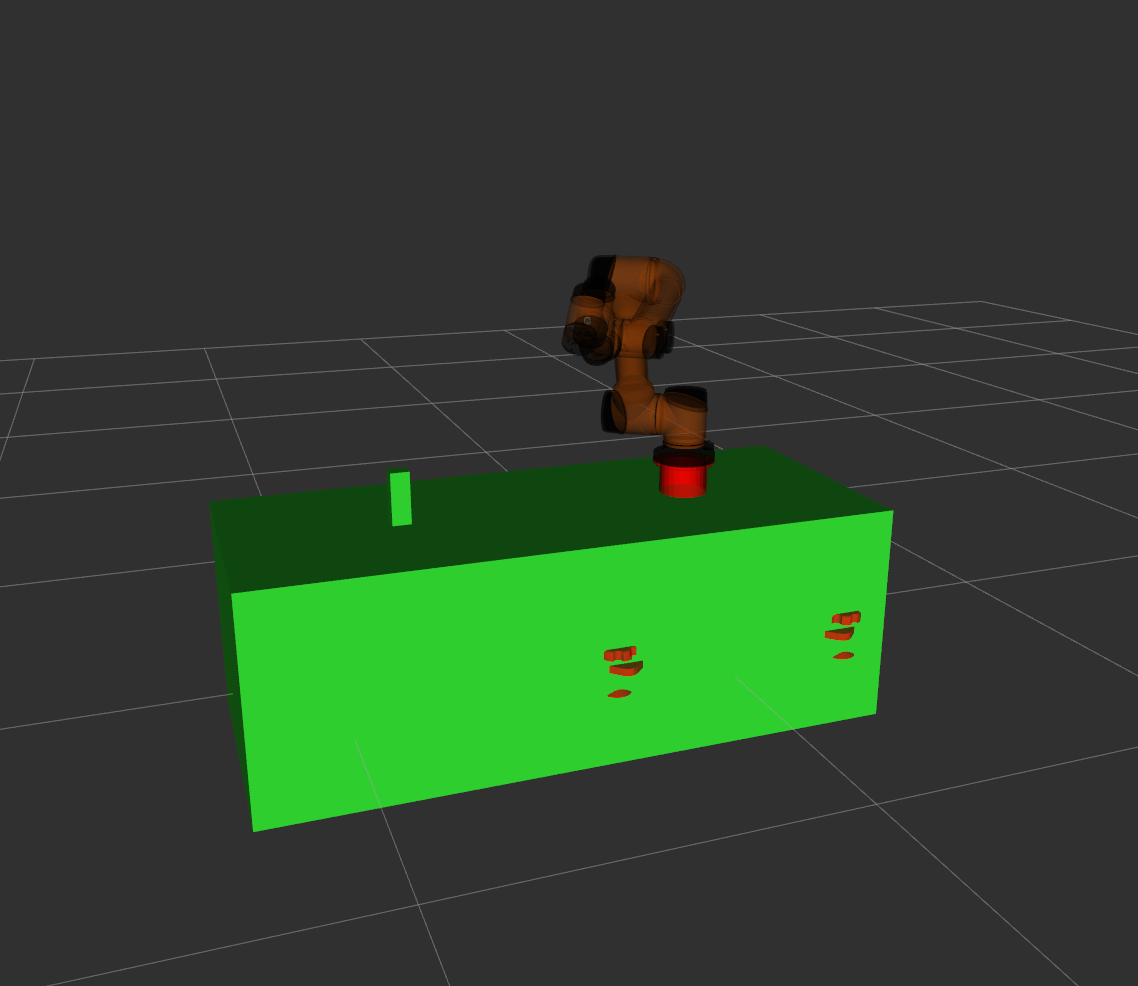}
  \end{subfigure}
  \begin{subfigure}[b]{0.43\linewidth}
    \centering
    \includegraphics[width=3.8cm,height=3cm]{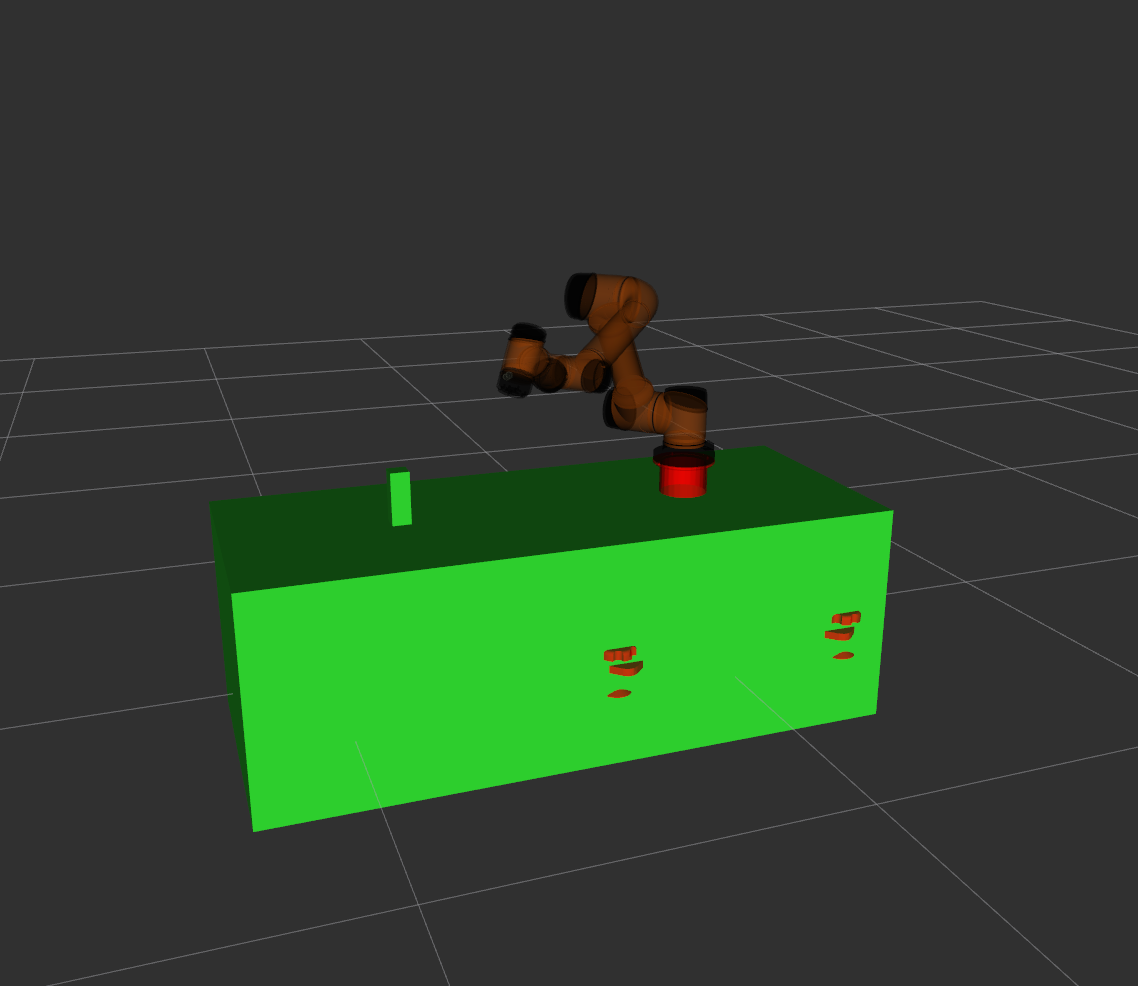}
  \end{subfigure}
  \begin{subfigure}[b]{0.43\linewidth}
    \centering
    \includegraphics[width=3.8cm,height=3cm]{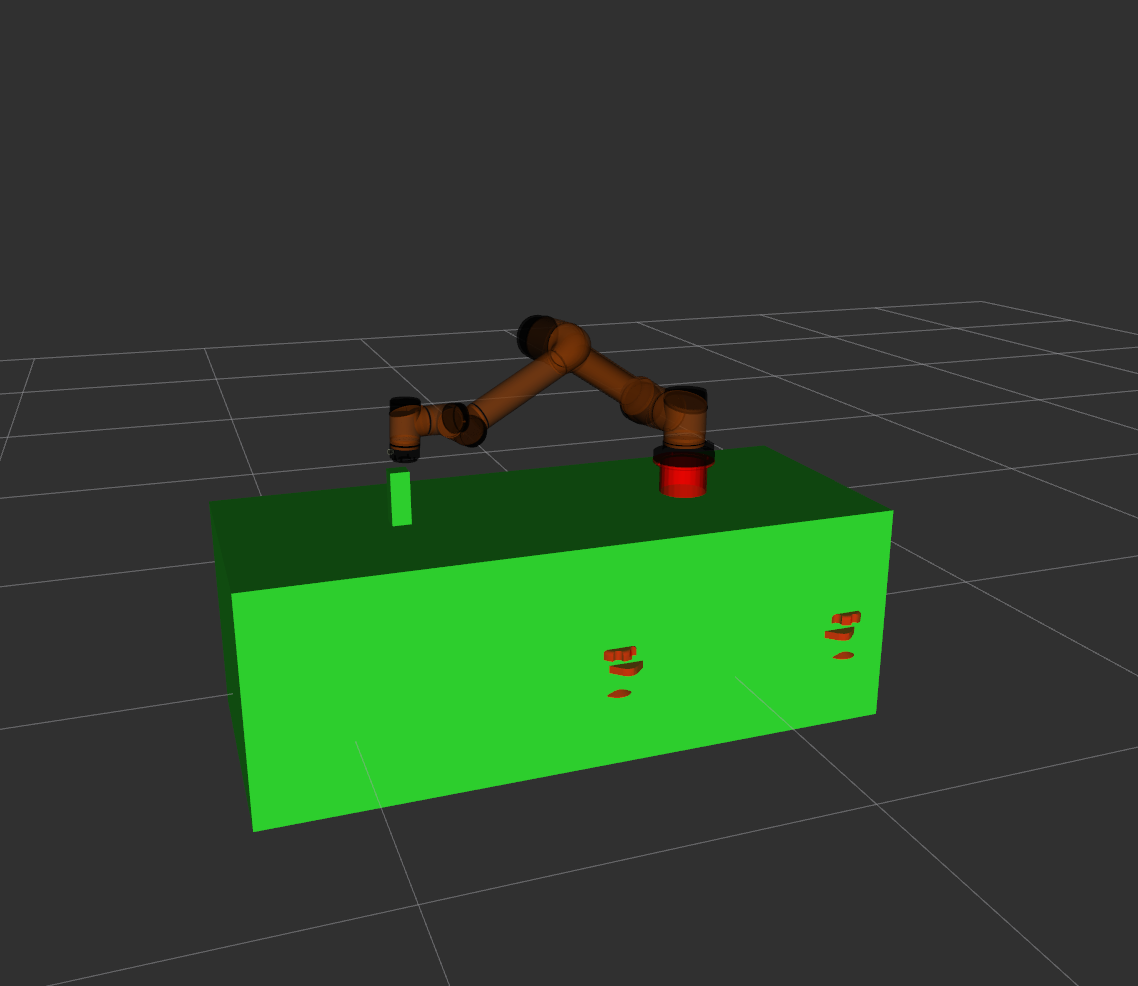}
  \end{subfigure}
  \caption{In a simulated experiment, the AuboReach environment performs a task using the best policy learned via GA+DDPG+HER.}
  \label{fig:auboReachEnvSim}
\end{figure}

\textit{FetchPick\&Place-v1}, \textit{FetchPush-v1}, \textit{FetchReach-v1},  \textit{FetchSlide-v1}, and \textit{DoorOpening} are the environments used to test robot learning on five different simulations tasks (see Figures \ref{finalPlots} and \ref{finalPlotsDoorOpening}). Figures \ref{fig:auboReachEnv} and \ref{fig:auboReachEnvSim} depict \textit{AuboReach} habitats, which are used in both simulated and real-world research. On these six gym situations, we test our algorithm. \textit{FetchPick\&Place}, \textit{FetchPush}, \textit{FetchReach}, and \textit{FetchSlide} are fetch environments from \cite{brockman2016openai}.
We created \textit{DoorOpening} and \textit{AuboReach}, which are custom-built gym environments. The following are the details of the six tasks:
\begin{itemize}
  \item \textbf{FetchPick\&Place}: The agent picks up the box from a table and moves to the goal position, which may be anywhere on the table or the area above it.
  \item \textbf{FetchPush}: In front of the agent, a box is stored. It pushes or rolls the box to the table's goal location. The agent is not permitted to take possession of the box.
  \item \textbf{FetchReach}: In the area around the end-effector, the agent must move it to the goal position.
  \item \textbf{FetchSlide}: The puck is placed on a slippery surface within reach of the agent. It must strike the puck with sufficient force that it (the puck) comes to a halt in front of the goal owing to friction.
  \item \textbf{DoorOpening}: A simulated Aubo i5 manipulator is positioned within reach of a door, with the door handle pointing in the direction of the robot. The goal is to force the door open by applying force to the area of the door handle. 
  \item \textbf{AuboReach}: An Aubo i5 manipulator, simulated or real, learns to achieve a desired joint configuration and pick up an object with a gripper.
\end{itemize}

\subsection{Running GA}

We ran the GA independently on each of these scenarios to test the effectiveness of our method and compared results to the hyperparameters' original values. We would anticipate that when comparing GA+DDPG+HER with DDPG+HER, PPO, A2C, and DDPG, the algorithm with the fewest episodes, running time, steps, and average epochs will do the best. This would illustrate why optimizing hyperparameters is crucial as opposed to using the algorithms' built-in default hyperparameters. Figure \ref{fig:compareFetchPusha} depicts the outcome of our \textit{FetchPush-v1} experiment. We used the GA to run the system and discover the best values for the hyperparameters $\tau$ and $\gamma$. We display results from ten GA runs because the GA is probabilistic, and the results show that optimized hyperparameters determined by the GA can lead to greater performance. The learning agent can learn faster and achieve a higher success rate.
Figure \ref{fig:compareFetchPushb} depicts one learning run for the initial hyperparameter set, as well as the average learning for the ten GA iterations. As we evaluated the genetic algorithm, the results displayed in figure \ref{fig:compareFetchPush} demonstrate changes when only two hyperparameters are tuned. We can see the potential for performance improvement. Our findings from optimizing all five hyperparameters back up our optimism, and they are detailed below.

Figure \ref{fig:compareFetchSlide} (b) shows a comparison of one original experiment with two averaged runs for optimizing hyperparameters $\tau$ and $\gamma$. Because this operation can take several hours to complete in a single run, and because it was a component of one of our initial tests, we only performed it twice. As we evaluated the genetic algorithm, the results displayed in Figures \ref{fig:compareFetchPush} and \ref{fig:compareFetchSlide} demonstrate changes when only two hyperparameters are tuned. We can see the potential for performance improvement. Our findings from optimizing all five hyperparameters back up our optimism, and they are detailed below.

After that, GA was used to optimize all hyperparameters, with the results presented in Figures \ref{finalPlots} and \ref{finalPlotsDoorOpening} for each task. Table \ref{table:drl1} compares the GA-discovered hyperparameters to the RL algorithm's initial hyperparameters. All the simulated environments, with the exception of AuboReach, used the same set of hyperparameters discovered by GA+DDPG+HER. As seen in table \ref{table:drl1}, yet another set of hyperparameters was generated for AuboReach. The learning rates, $\alpha_{actor}$ and $\alpha_{critic}$ are the same as they were in the beginning, while the other four hyperparameters have different values. Figures \ref{finalPlots} and \ref{finalPlotsDoorOpening} indicate that the GA-discovered hyperparameters outperformed the original hyperparameters, suggesting that the learning agent was able to learn more quickly. All of the plots in the previous figure have been averaged over ten runs.

\begin{table}[ht]
\centering
\begin{tabular}{|p{0.15\linewidth}|p{0.15\linewidth}|p{0.15\linewidth}|p{0.15\linewidth}|p{0.15\linewidth}|} 
 \hline
 {} & {} & \textbf{All environments except Aubo-i5} & \textbf{Aubo-i5 - Fixed Initial and Target state} & \textbf{Aubo-i5 - Random Initial and Target state} \\
 \hline
 \textbf{hyper- parameters} & \textbf{DDPG+ HER} & \textbf{GA+ DDPG+ HER} & \textbf{GA+ DDPG+ HER} & \textbf{GA+ DDPG+ HER}\\ 
 \hline
 $\gamma$ & 0.98 & 0.928 & 0.949 & 0.988\\ 
 \hline
 $\tau$ & 0.95 & 0.484 & 0.924 & 0.924\\
 \hline
 $\alpha_{actor}$ & 0.001 & 0.001 & 0.001 & 0.001 \\
 \hline
 $\alpha_{critic}$ & 0.001 & 0.001 & 0.001 & 0.001 \\
 \hline
 $\epsilon$ & 0.3 & 0.1 & 0.584 & 0.912\\ 
 \hline
 $\eta$ & 0.2 & 0.597 & 0.232 & 0.748\\
 \hline
\end{tabular}
\caption{DDPG+HER vs. GA+DDPG+HER values of hyperparameters.}
\label{table:drl1}
\end{table}

Figures \ref{fig:auboReachEnv} and \ref{fig:auboReachEnvSim} show how GA+DDPG+HER settings (as listed in table \ref{table:drl1}) were applied to a custom-built gym environment for the Aubo-i5 robotic manipulator. The motors in this environment are controlled by the \textit{MOVEit} package \cite{hernandez2017design} , but the DDPG+HER works as a brain for movement. Each action (series of joint states) that the robot should take is determined by DDPG+HER. The results were at first unexpected. Each epoch took several hours to complete ($>$ 10-15 hours). We did not complete the entire curriculum because it could take several weeks. The same holds for DDPG+HER settings. This is because the movement speed of the Aubo i5 robotic manipulator was kept sluggish in both simulation and real-world studies to avoid any unexpected sudden movements, which could result in harm.
In the \textit{AuboReach} setting, there were also planning and execution processes involved in the successful completion of each action. \textit{AuboReach}, unlike the other gym environments covered in this study, could only run on a single CPU. This is because other environments were introduced in MuJoCo and could easily run with the maximum number of CPUs available. MuJoCo can create several instances for training, which allows for faster learning. AuboReach must only perform one action at a time, much like a real robot. Because of these characteristics, training in this setting takes a long period.

\subsection{Modifications required for AuboReach}

The GA+DDPG+HER settings were then applied to the \textit{AuboReach} environment, but only to action values. This means that in both simulated and real-world studies, the robot did not run to do the action. This further means that whatever decision the DDPG+HER makes, the robot is taken to have made the decision. This is further supported by the fact that \textit{MOVEit} handles the hassle of communicating move signals to numerous joints. This is accurate since the robot can do each action determined by the DDPG+HER algorithm. This is due to the planning and execution required for internal joint movement in the robot \cite{field1996iterative}. This also eliminates the possibility of a collision, which may have occurred if these procedures had been skipped. Each epoch now took less than a minute to complete, a considerable reduction in training time that allows this environment to be used for training.

The GA+DDPG+HER settings (table \ref{table:drl1}) were re-applied now that some of the environmental challenges had been overcome. These hyperparameters did not outperform the original hyperparameters in any way. We believe this is due to the fact that this environment is far more intricate and unique than others. We took into account even more aspects to ensure that the environment is trainable. For training and testing, this environment employs four joints (instead of six). \textit{shoulder}, \textit{forearm}, \textit{upper-arm}, and \textit{wrist1} are the ones used. This was done to ensure that the learning could be finished in a reasonable amount of time. Each joint has a range of -1.7 to 1.7 radians. The robot's initial and reset states were both set to upright, i.e., [0, 0, 0, 0].

\begin{figure}
\centering
  \begin{subfigure}[b]{\linewidth}
    \centering
    \includegraphics[width=5.5cm,height=4cm]{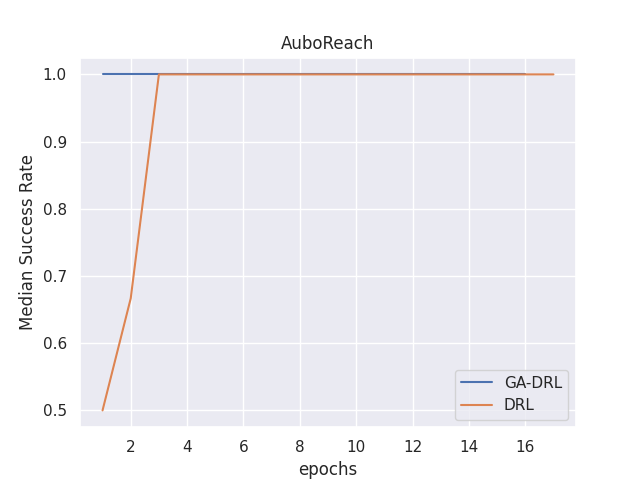}
    \caption{Trained initial and target state}
    \label{fig:compareAuboReach}
  \end{subfigure}
  \begin{subfigure}[b]{\linewidth}
    \centering
    \includegraphics[width=5.5cm,height=4cm]{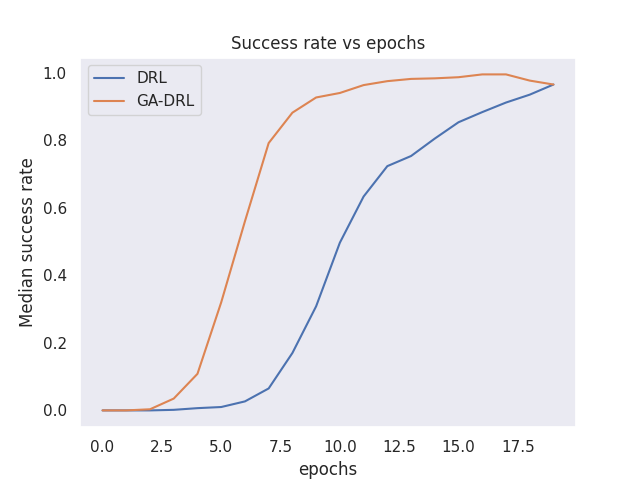}
    \caption{Training is done with random initial and target joint states with 1 CPU}
    \label{fig:compareAuboReachRandomInitTarget}
  \end{subfigure}
  \caption{The \textit{AuboReach} task's success rate against epochs. This graph represents the average of over ten runs.}
  \label{fig:compareAuboReachComplete}
\end{figure}

The GA+DDPG+HER method was tweaked for better learning and faster hyperparameter search, in addition to minor adjustments to the environment. The definition of success was changed to include ten successful epochs. This means that the GA was regarded successful if it had a 100\% success rate for ten consecutive epochs. When  $\alpha_{actor}$ and $\alpha_{critic}$ are both greater than 0.001, learning never converges, according to experiments. As a result, the values of $\alpha_{actor}$ and $\alpha_{critic}$ were set to 0.001. Multi-threading can occur when only action values are employed; hence four CPUs could be utilized. If the cumulative discrepancy between the target and the achieved joint states is less than 0.1 radians, AuboReach considers the DDPG+HER-determined joint states to be a success. [-0.503, 0.605, -1.676, 1.391] were chosen as the objective joint states.
We were able to find a new set of hyperparameters using these changes to the algorithm, as shown in Table \ref{table:drl1}. Figure \ref{fig:compareAuboReach} shows the difference in success rates between DDPG+HER and GA+DDPG+HER during training. Without a doubt, the GA+DDPG+HER outperforms the DDPG+HER.

After the GA+DDPG+HER had determined the best hyperparameters for the \textit{AuboReach} environment, the training was repeated using four CPUs to determine the best policy. The robots were then subjected to this policy in both simulated and real-world testing. The crucial point to remember here is that for testing purposes, CPU consumption was reduced to one. In both studies, the robot was able to transition from the training's beginning joint space to the goal joint space. Because unpredictability was not added during training, the environment was constrained to only one possible path. Since both DDPG+HER and GA+DDPG+HER eventually achieved a 100\% success rate, there was no discernible difference throughout testing. The main distinction is the speed with which the environment may learn given a set of hyperparameters.

The \textit{AuboReach} environment was updated in another experiment to train on random joint states. The robot may now start and reach objectives in various joint states during testing thanks to the update. The GA was run on this environment, and the hyperparameters discovered by the GA are shown in table \ref{table:drl1}. Figure \ref{fig:compareAuboReachRandomInitTarget} shows that the plot of GA+DDPG+HER is still better than DDPG+HER. Figures \ref{fig:auboReachEnv} and \ref{fig:auboReachEnvSim} show the robot in action as it accomplishes the task of picking the object in real and simulated tests.

The use of GA+DDPG+HER, in the \textit{AuboReach} environment resulted in automatic DDPG+HER hyperparameter adjustment, which improved the algorithm's performance.

\subsection{Training evaluation}

Monitoring the status of a GA while it is operating is critical for optimizing the system's performance. Because of the way GAs work, certain chromosomes will perform better than others. As a result, the performance graph is unlikely to be a smooth rise curve. Some fitness function assessments result in a zero, implying that the chromosome is unsuited for use. Despite the non-smooth slope, it is believed that the system's overall performance would improve as GA advances.

\begin{figure}[!h]
\centering
\begin{multicols}{2}
    \includegraphics[width=5cm,height=4cm]{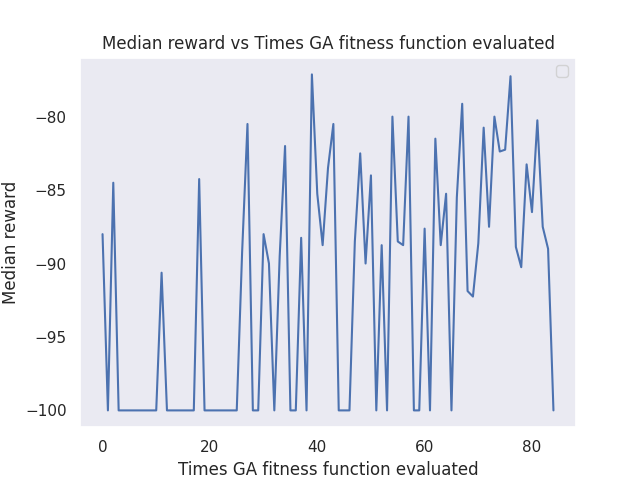}
    \subcaption{FetchPick\&Place - Median reward vs Times GA fitness function evaluated}
    \includegraphics[width=5cm,height=4cm]{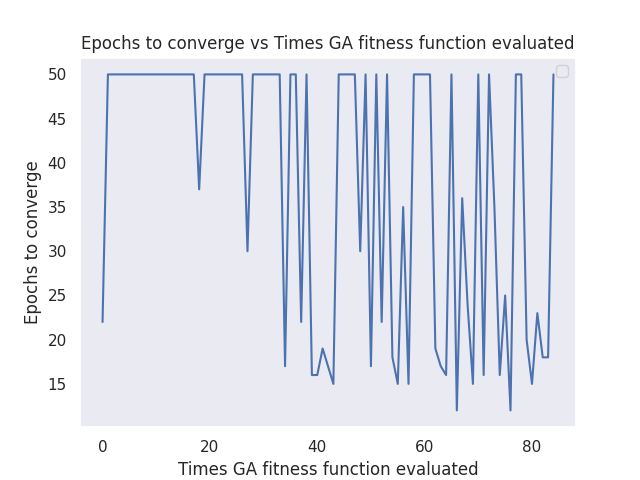}
    \subcaption{FetchPick\&Place - Epochs vs Times GA fitness function evaluated}
    \includegraphics[width=5cm,height=4cm]{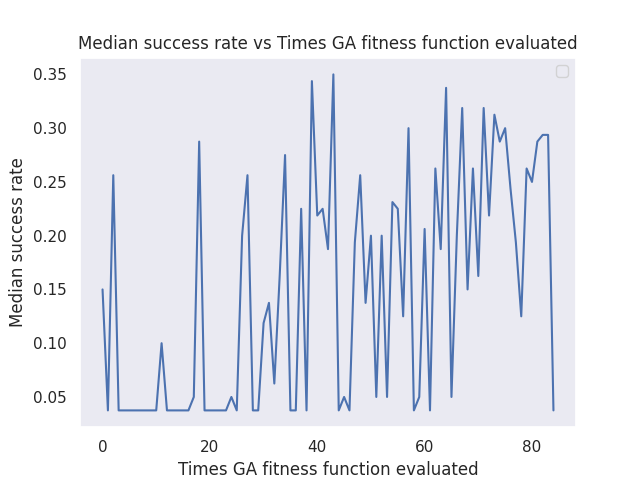}
    \subcaption{FetchPick\&Place - Median success rate vs Times GA fitness function evaluated}
    \includegraphics[width=5cm,height=4cm]{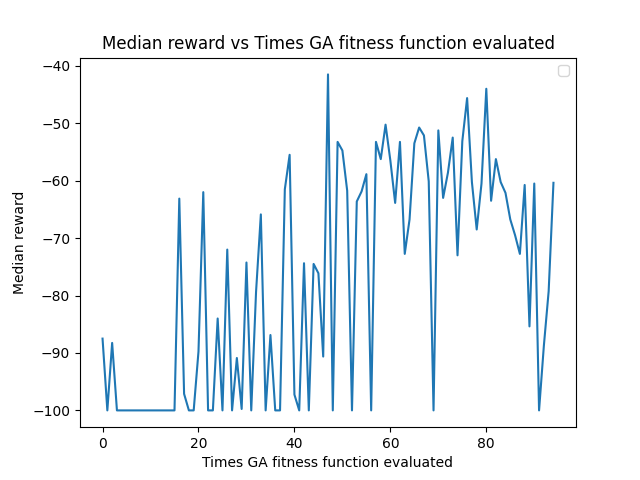}
    \subcaption{FetchPush - Median reward vs Times GA fitness function evaluated}
    \includegraphics[width=5cm,height=4cm]{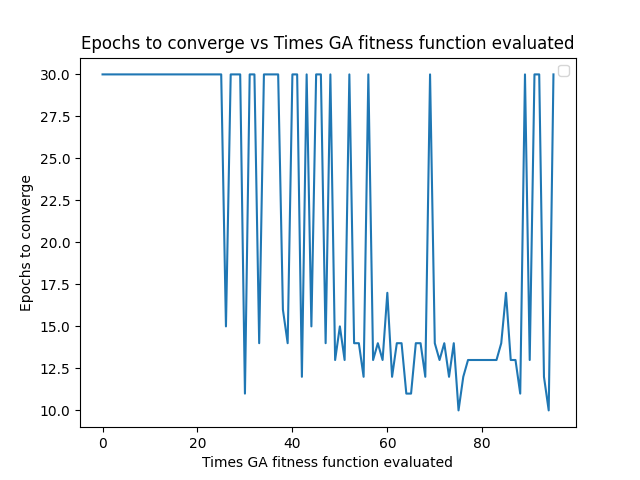}
    \subcaption{FetchPush - Epochs vs Times GA fitness function evaluated}
    \includegraphics[width=5cm,height=4cm]{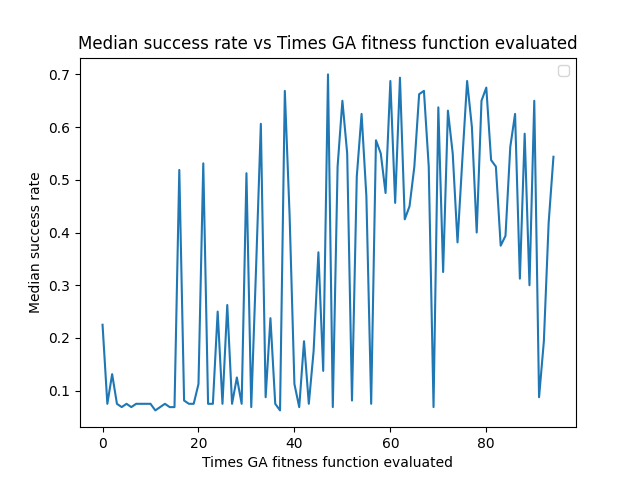}
    \subcaption{FetchPush - Median success rate vs Times GA fitness function evaluated}
\end{multicols}
\caption{When GA finds all six hyperparameters, the GA+DDPG+HER training evaluation charts. This is the outcome of a single GA run.}
\label{fig:gaTrainingEvaluationPlots}
\end{figure}

\begin{figure}[!h]
\centering
\begin{multicols}{2}
    \includegraphics[width=5cm,height=4cm]{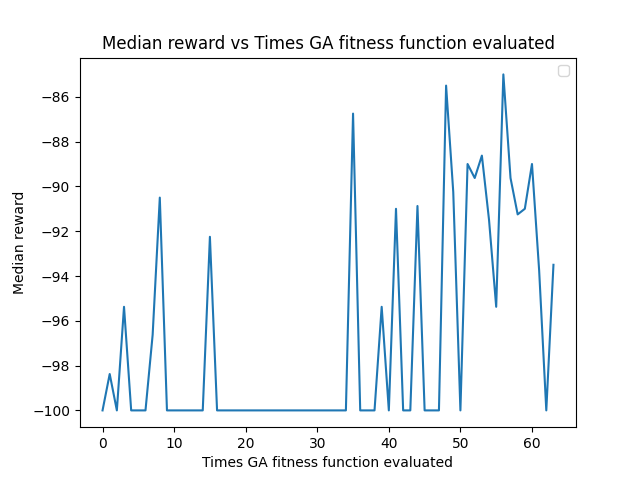}
    \subcaption{FetchSlide - Median reward vs Times GA fitness function evaluated}
    \includegraphics[width=5cm,height=4cm]{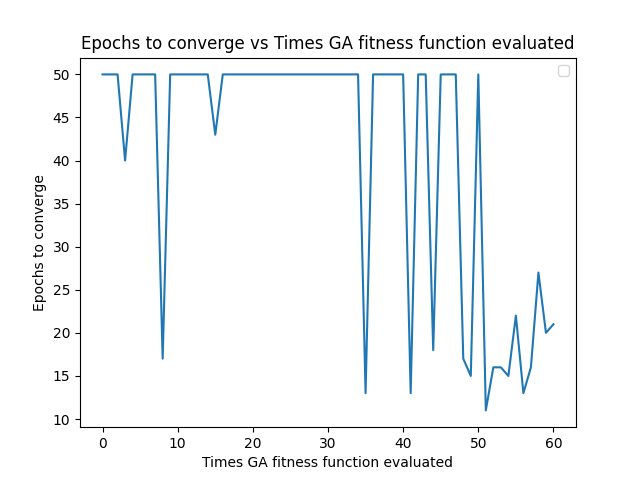}
    \subcaption{FetchSlide - Epochs vs Times GA fitness function evaluated}
    \includegraphics[width=5cm,height=4cm]{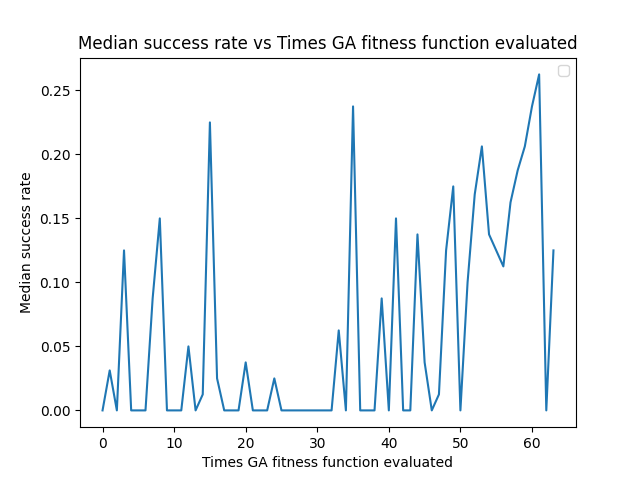}
    \subcaption{FetchSlide - Median success rate vs Times GA fitness function evaluated}
    \includegraphics[width=5cm,height=4cm]{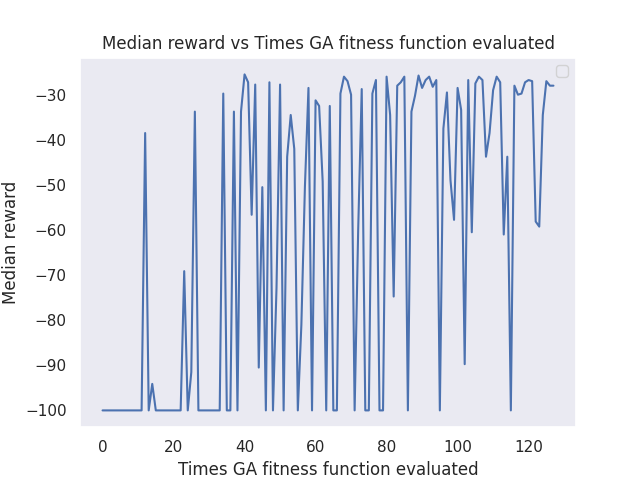}
    \subcaption{DoorOpening - Median reward vs Times GA fitness function evaluated}
    \includegraphics[width=5cm,height=4cm]{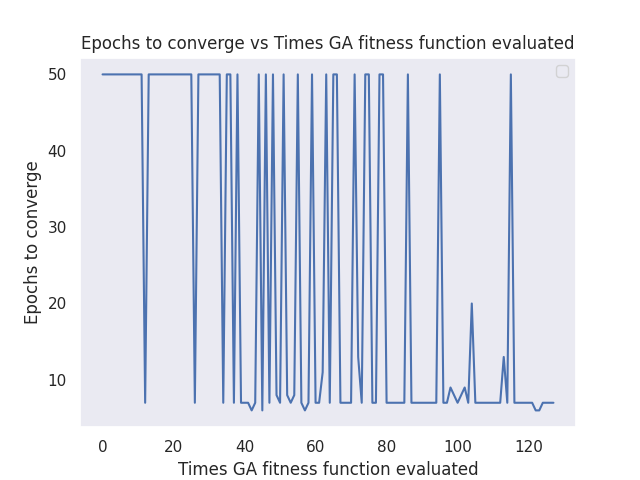}
    \subcaption{DoorOpening - Epochs vs Times GA fitness function evaluated}
    \includegraphics[width=5cm,height=4cm]{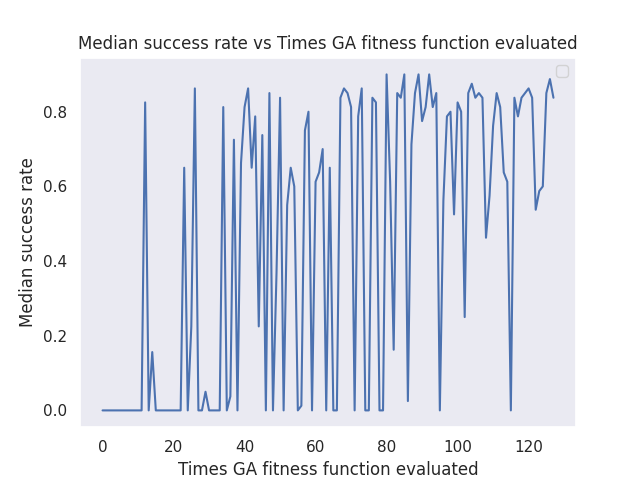}
    \subcaption{DoorOpening - Median success rate vs Times GA fitness function evaluated}
\end{multicols}
\caption{When all six hyperparameters are identified by GA, the GA+DDPG+HER training evaluation plots. One GA run yielded this result.}
\label{fig:gaTrainingEvaluationPlotAuboReach}
\end{figure}

\begin{figure}[!h]
\centering
\begin{multicols}{2}
    \includegraphics[width=5cm,height=4cm]{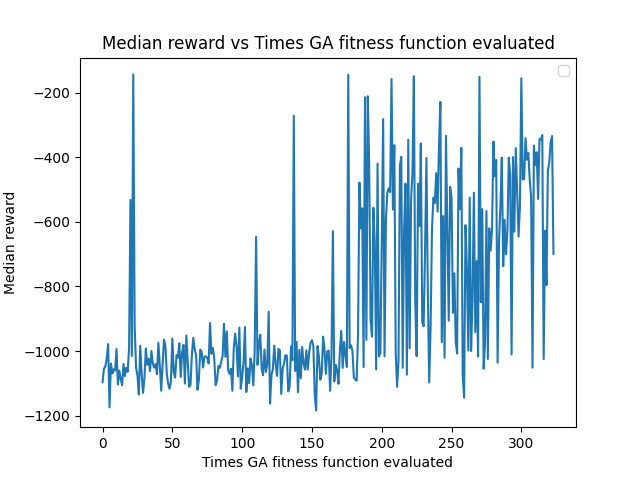}
    \subcaption{AuboReach - Median reward vs Times GA fitness function evaluated}
    \includegraphics[width=5cm,height=4cm]{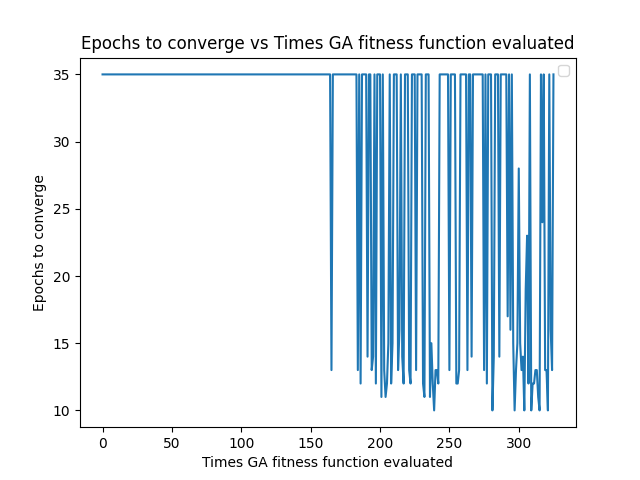}
    \subcaption{AuboReach - Epochs vs Times GA fitness function evaluated}
    \includegraphics[width=5cm,height=4cm]{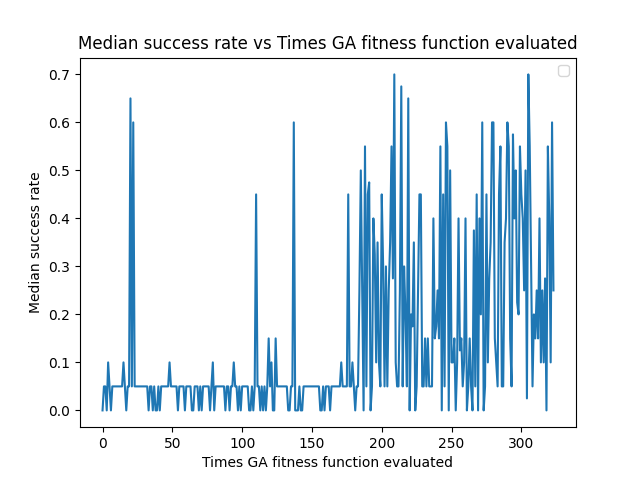}
    \subcaption{AuboReach - Median success rate vs Times GA fitness function evaluated}
    \includegraphics[width=5cm,height=4cm]{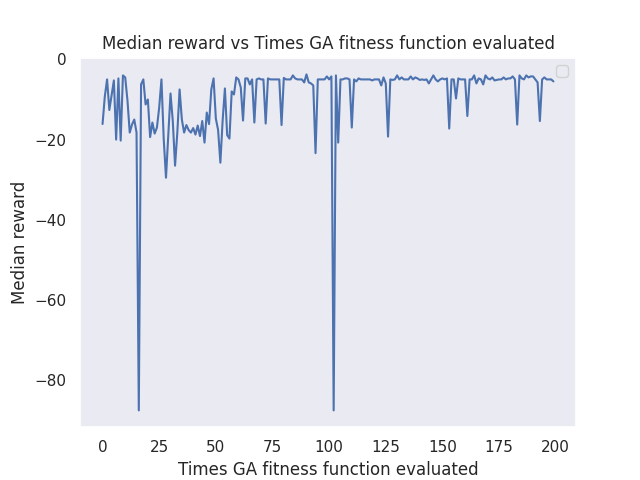}
    \subcaption{FetchReach - Median reward vs Times GA fitness function evaluated}
    \includegraphics[width=5cm,height=4cm]{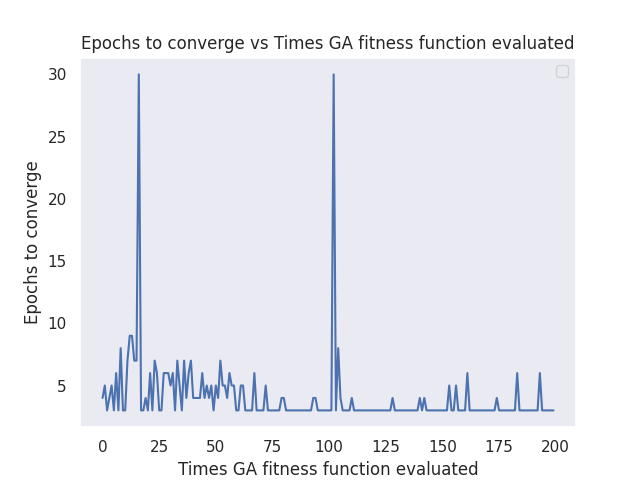}
    \subcaption{FetchReach - Epochs vs Times GA fitness function evaluated}
    \includegraphics[width=5cm,height=4cm]{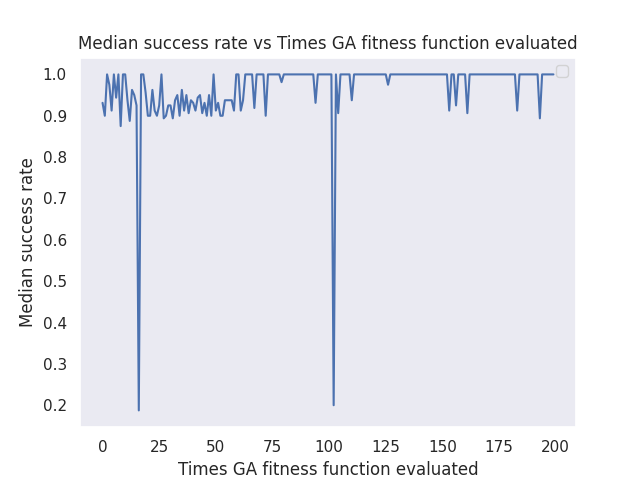}
    \subcaption{FetchReach - Median success rate vs Times GA fitness function evaluated}
\end{multicols}
\caption{When GA finds all six hyperparameters, the GA+DDPG+HER training evaluation charts. This is the outcome of a single GA run.}
\label{fig:gaTrainingEvaluationPlotAuboReach2}
\end{figure}

We created numerous charts to track GA's progress while we looked for the best hyperparameters. Figures \ref{fig:gaTrainingEvaluationPlots}, \ref{fig:gaTrainingEvaluationPlotAuboReach}, and \ref{fig:gaTrainingEvaluationPlotAuboReach2} show how the system's performance improves as GA progresses. Median success rate across fitness function evaluations, total reward across episodes, and epochs to accomplish the target throughout fitness function evaluations are all hyperparameters taken into account when evaluating training performance. It can be seen that the system's overall performance is improving. With fitness function evaluations, the overall reward increases as the agent's episodes and epochs to accomplish the goal decrease. This indicates that the GA is on the right road in terms of determining the best hyperparameter values.
We plotted data for only one GA run and limited the duration of a GA run because each GA run requires several hours to many days of run time. When we started to see results, we decided to terminate the GA.

Now that the GA has behaved as intended, we will evaluate the system's efficiency using the GA's discovered hyperparameters.

\subsection{Efficiency evaluation}

\begin{figure}[!h]
\centering
\begin{multicols}{2}
    \includegraphics[width=5cm,height=4cm]{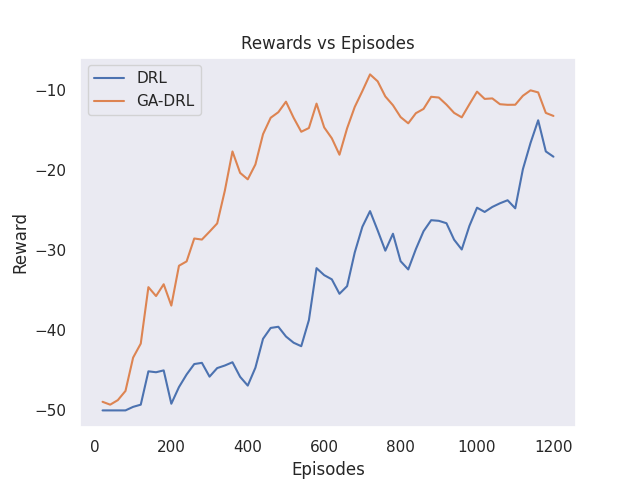}
    \subcaption{FetchPick\&Place}
    \includegraphics[width=5cm,height=4cm]{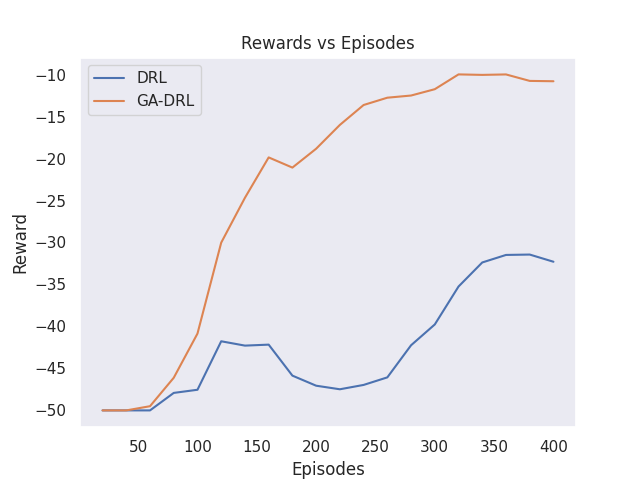}
    \subcaption{FetchPush}
    \includegraphics[width=5cm,height=4cm]{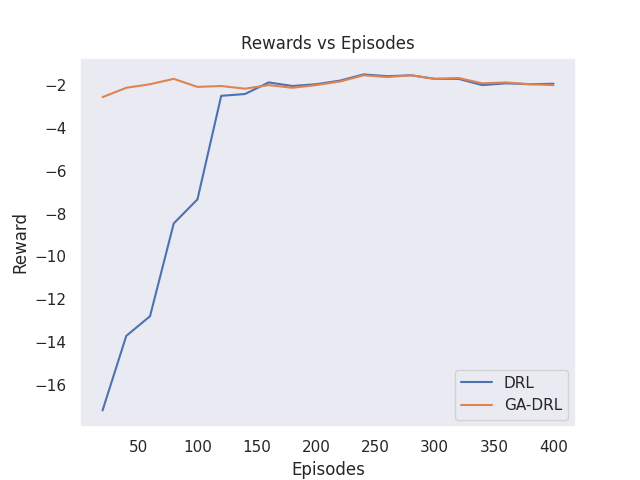}
    \subcaption{FetchReach}
    \includegraphics[width=5cm,height=4cm]{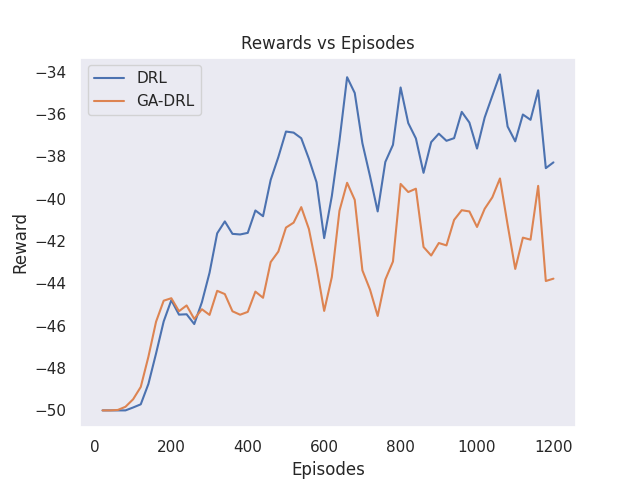}
    \subcaption{FetchSlide}
    \includegraphics[width=5cm,height=4cm]{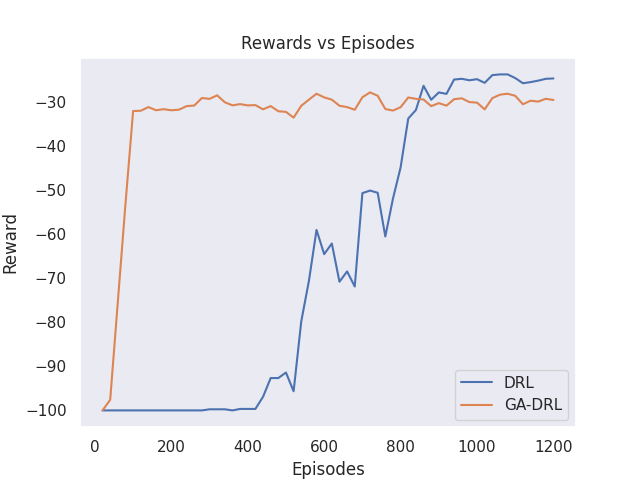}
    \subcaption{DoorOpening}
    \includegraphics[width=5cm,height=4cm]{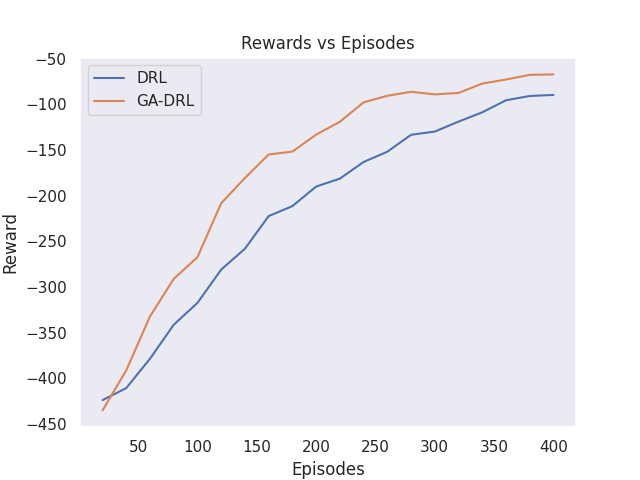}
    \subcaption{AuboReach}
\end{multicols}
\caption{When all six hyperparameters are determined using GA, the DDPG+HER vs. GA+DDPG+HER efficiency evaluation plots (Total reward vs episodes). Over ten runs, all graphs are averaged. }
\label{fig:analysisPlots}
\end{figure}

We generated data for various hyperparameters to evaluate and compare the efficiency of the GA+DDPG+HER algorithm for training the agent to accomplish a task. These hyperparameters are good indicators of the algorithm's efficiency. For the majority of the training tasks, as shown in Figure \ref{fig:analysisPlots}, the total reward has increased significantly. The DDPG+HER algorithm's efficiency was significantly improved when rewards were raised. Because it is steered considerably faster towards the desired task, the agent can learn a lot faster. To get an unbiased review, we averaged these plots over ten runs. With GA+DDPG+HER, the \textit{FetchSlide} environment performed worse. We believe that this is due to the task's complexity. The maximum number used for that hyperparameter was used to represent in the tables the tasks that did not attain the target during training.

We also generated more data to evaluate the episodes, running time (s), steps, and epochs that an agent must learn to achieve the desired goal. Tables \ref{table:averageepisodes}-\ref{table:averageepochs} present this information. The information in the tables is an average of ten runs. Table \ref{table:averageepisodes} compares the number of episodes an agent needs to achieve a goal. The bolded values imply superior performance, and the majority of environments outperform the rest of the algorithms. The task is learned in $54.34\%$ fewer episodes in the \textit{FetchPush} environment than in DDPG+HER.

\begin{table}[ht]
\centering
\begin{tabular}{|p{0.1\linewidth}|p{0.1\linewidth}|p{0.1\linewidth}|p{0.1\linewidth}|p{0.1\linewidth}|p{0.1\linewidth}|p{0.1\linewidth}|} 
 \hline
 \textbf{Method} & \textbf{Fetch Pick \& Place} & \textbf{Fetch Push} & \textbf{Fetch Reach} & \textbf{Fetch Slide} & \textbf{Door Opening} & \textbf{Aubo Reach}\\ 
 \hline
 DDPG+ HER & 6,000 & 2,760 & 100 & \textbf{4,380} & 960 & 320\\ 
 \hline
 GA+ DDPG+ HER & \textbf{2,270} & \textbf{1,260} & \textbf{60} & 6,000 & \textbf{180} & \textbf{228}\\
 \hline
 PPO & 2,900 & 2,900 & 1,711 & 4,880 & 1,500 & 2,900\\
 \hline
 A2C & 119,999 & 119,999 & 119,999 & 119,999 & 119,999 & 32,512.3\\
 \hline
 DDPG & 10,000 & 2,000 & 423 & 10,000 & 706 & 1,000\\
 \hline
\end{tabular}
\caption{Efficiency evaluation: For all activities, compare average (over ten runs) episodes to accomplish the target.}
\label{table:averageepisodes}
\end{table}

Another factor to consider while evaluating an algorithm's efficiency is its running time. Seconds are used to measure time. The algorithm is better if it takes less time to learn the task. In the majority of the environments, the GA+DDPG+HER algorithm had the lowest running time, as shown in Table \ref{table:averagerunningtime}. For example, When compared to the DDPG+HER algorithm, \textit{FetchPush} with GA+DDPG+HER takes about $57.004\%$ less time.

\begin{table}[ht]
\centering
\begin{tabular}{|p{0.1\linewidth}|p{0.1\linewidth}|p{0.1\linewidth}|p{0.1\linewidth}|p{0.1\linewidth}|p{0.1\linewidth}|p{0.1\linewidth}|} 
 \hline
 \textbf{Method} & \textbf{Fetch Pick \& Place} & \textbf{Fetch Push} & \textbf{Fetch Reach} & \textbf{Fetch Slide} & \textbf{Door Opening} & \textbf{Aubo Reach}\\ 
 \hline
 DDPG+ HER & 3,069.981 & 1,314.477 & 47.223 & \textbf{2,012.645} & 897.816 & 93.258\\ 
 \hline
 GA+ DDPG+ HER & \textbf{1,224.697} & \textbf{565.178} & \textbf{28.028} & 3,063.599 & \textbf{167.883} & \textbf{66.818}\\
 \hline
 PPO & 1,964.411 & 2,154.052 & 776.512 & 2,379.393 & 997.779 & 710.576\\
 \hline
 A2C & 2,025.344 & 2,082.807 & 2,061.807 & 2,268.114 & 2,718.769 & 214.075\\
 \hline
 DDPG & 5,294.984 & 1,000.586 & 236.4 & 5,346.516 & 438.7 & 1,721.992\\
 \hline
\end{tabular}
\caption{Efficiency evaluation: For all activities, compare the average (over ten runs) running time (s) to attain the target.}
\label{table:averagerunningtime}
\end{table}

The average number of steps necessary to achieve the goal is another factor to consider when analyzing and investigating the GA+DDPG+HER algorithm's performance. The average number of steps taken by an agent in each environment is shown in Table \ref{table:averagesteps}. Except for the \textit{FetchSlide} environment, most of the environments, when employed with GA+DDPG+HER, outperform all other algorithms. When compared to the DDPG+HER algorithm,  \textit{FetchPush} with GA+DDPG+HER takes about $54.35\%$ fewer steps.

\begin{table}[ht]
\centering
\begin{tabular}{|p{0.1\linewidth}|p{0.1\linewidth}|p{0.1\linewidth}|p{0.1\linewidth}|p{0.1\linewidth}|p{0.1\linewidth}|p{0.1\linewidth}|} 
 \hline
 \textbf{Method} & \textbf{Fetch Pick \& Place} & \textbf{Fetch Push} & \textbf{Fetch Reach} & \textbf{Fetch Slide} & \textbf{Door Opening} & \textbf{Aubo Reach}\\ 
 \hline
 DDPG+ HER & 300,000 & 138,000 & 5,000 & \textbf{219,000} & 48,000 & 65,600\\ 
 \hline
 GA+ DDPG+ HER & \textbf{113,000} & \textbf{63,000} & \textbf{3,000} & 300,000 & \textbf{9,000} & \textbf{46,000}\\
 \hline
 PPO & 595,968 & 595,968 & 324,961 & 1,000,000 & 301,056 & 595,968\\
 \hline
 A2C & 600,000 & 600,000 & 600,000 & 600,000 & 600,000 & 162,566.5\\
 \hline
 DDPG & 500,000 & 100,000 & 21,150 & 500,000 & 35,300 & 200,000\\
 \hline
\end{tabular}
\caption{Efficiency evaluation: For all activities, compare the average (over ten runs) steps taken to attain the target.}
\label{table:averagesteps}
\end{table}

The number of epochs taken by the agent to attain the goal is the final hyperparameter used to compare the competency of GA+DDPG+HER with four other algorithms. The average epochs for all of the environments are shown in Table \ref{table:averageepochs}. Almost all environments outperform GA+DDPG+HER in terms of efficacy. \textit{FetchPush}, for example, uses $54.35\%$ fewer epochs with GA+DDPG+HER than it does with DDPG+HER. Following that, we give a comparison of the GA+DDPG+HER algorithm to the other algorithms.

\begin{table}[ht]
\centering
\begin{tabular}{|p{0.1\linewidth}|p{0.1\linewidth}|p{0.1\linewidth}|p{0.1\linewidth}|p{0.1\linewidth}|p{0.1\linewidth}|p{0.1\linewidth}|} 
 \hline
 \textbf{Method} & \textbf{Fetch Pick \& Place} & \textbf{Fetch Push} & \textbf{Fetch Reach} & \textbf{Fetch Slide} & \textbf{Door Opening} & \textbf{Aubo Reach}\\ 
 \hline
 DDPG+ HER & 60 & 27.6 & 5 & \textbf{43.8} & 47 & 16\\ 
 \hline
 GA+ DDPG+ HER & \textbf{22.6} & \textbf{12.6} & \textbf{3} & 60 & \textbf{8} & \textbf{11.4}\\
 \hline
 PPO & 290 & 290 & 171.1 & 488 & 150 & 290\\
 \hline
 A2C & 1,200 & 1,200 & 1,200 & 1,200 & 1,200 & 325.2\\
 \hline
 DDPG & 1,000 & 100 & 42.3 & 1,000 & 70.6 & 100\\
 \hline
\end{tabular}
\caption{Efficiency evaluation: Average (over ten runs) epochs comparison to reach the goal, for all the tasks.}
\label{table:averageepochs}
\end{table}

Next, we present the overall analysis of the GA+DDPG+HER algorithm compared to the other algorithms.

\subsection{Analysis}
\begin{figure}[!h]
\centering
\begin{multicols}{2}
    \includegraphics[width=5cm,height=4cm]{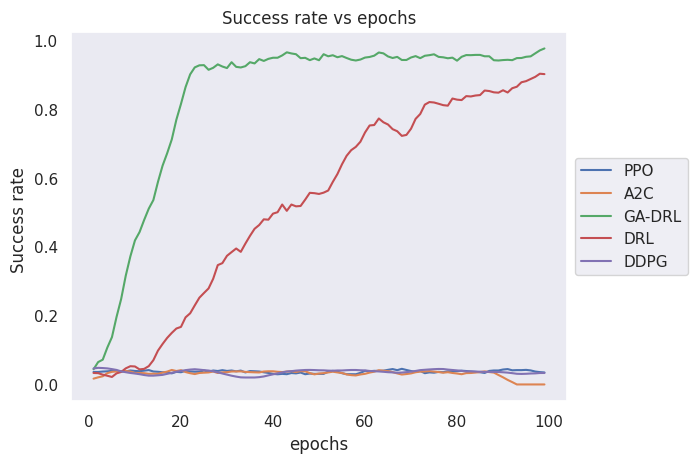}
    \subcaption{FetchPick\&Place}
    \includegraphics[width=5cm,height=4cm]{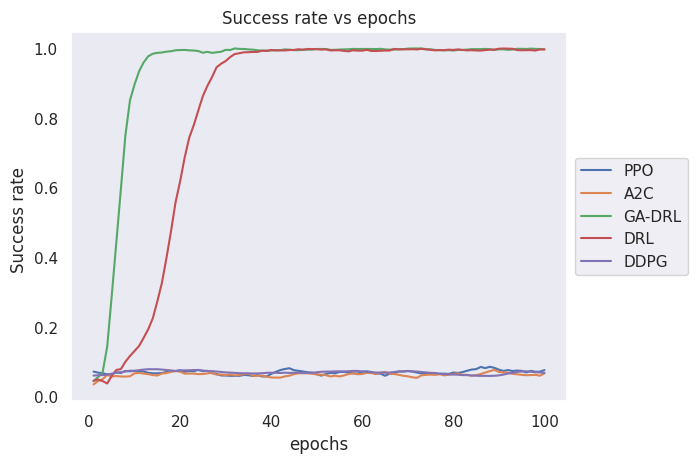}
    \subcaption{FetchPush}
    \includegraphics[width=5cm,height=4cm]{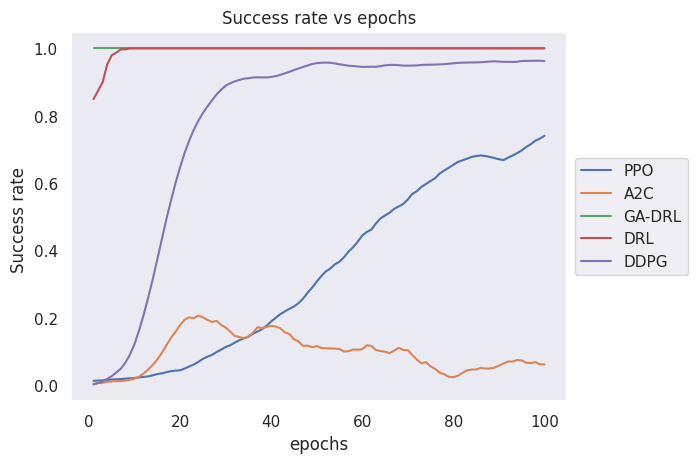}
    \subcaption{FetchReach}
    \includegraphics[width=5cm,height=4cm]{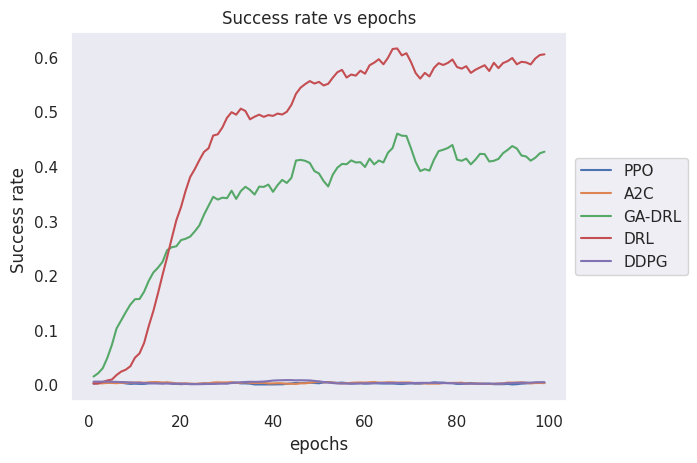}
    \subcaption{FetchSlide}
    \includegraphics[width=5cm,height=4cm]{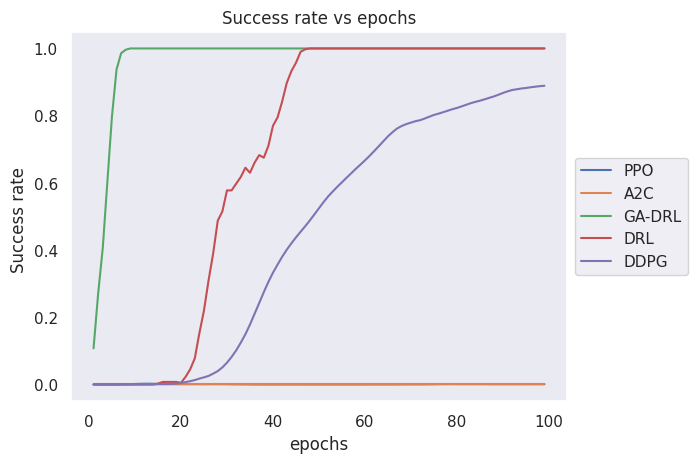}
    \subcaption{DoorOpening}
    \includegraphics[width=5cm,height=4cm]{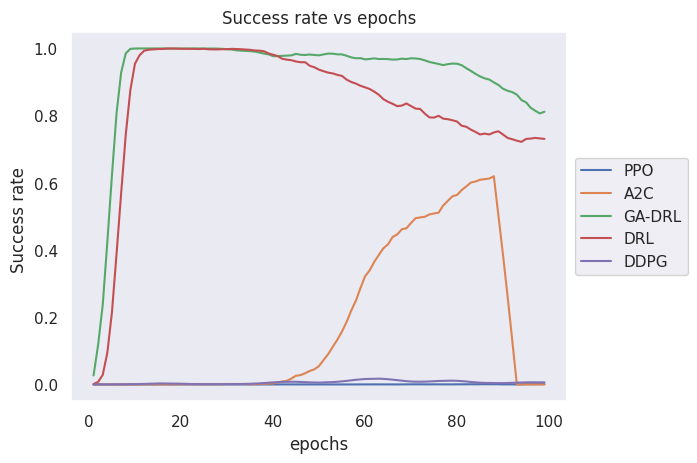}
    \subcaption{AuboReach}
\end{multicols}
\caption{GA+DDPG+HER comparison with PPO \cite{schulman2017proximal}, A2C \cite{mnih2016asynchronous}, DDPG+HER (DDPG\cite{lillicrap2015continuous} + HER\cite{andrychowicz2017hindsight}) and DDPG \cite{lillicrap2015continuous}. All plots are averaged over ten runs.}
\label{fig:fetchReachCompareAllMethods}
\end{figure}

We provided various findings and the mechanism for judging the efficacy of GA+DDPG+HER versus DDPG+HER, PPO, A2C, and DDPG algorithms in the previous sub-sections. Overall, GA+DDPG+HER works best, with one exception of the \textit{FetchSlide} environment. The average comparison tables illustrate that different values of the assessment hyperparameters can be assumed by each environment. This is determined by the type of task that the agent is attempting to learn. While the majority of the tasks outperformed DDPG+HER with more than a $50\%$ increase in efficiency, \textit{FetchSlide} underperformed DDPG+HER. The task's goal is also credited with this performance. The end-effector does not physically go to the desired position to place the box, which makes this task unique. GA+DDPG+HER was tested using a variety of hyperparameters and an average of over ten runs. This is sufficient proof that GA+DDPG+HER outperformed several algorithms. Figures \ref{finalPlots}, \ref{finalPlotsDoorOpening} and \ref{fig:compareAuboReachRandomInitTarget} support our claim by demonstrating that when GA+DDPG+HER is utilized, the task may be learned substantially faster in most environments. Finally, in Figure \ref{fig:fetchReachCompareAllMethods}, we compare the results of five different algorithms (PPO, A2C, DDPG, DDPG+HER, GA+DDPG+HER). As can be shown, GA+DDPG+HER outperforms all other algorithms in every environment with the exception of \textit{FetchSlide}, where DDPG+HER performs better than GA+DDPG+HER. Even for the winning DDPG+HER, we think the overall success rate on this problem (\textit{FetchSlide}) is low (the highest success rate is somewhere around 0.6 or $60\%$). In our opinion, GA+DDPG+HER works similarly to its counterpart DDPG+HER in terms of problem-solving speed but differs when an issue may have less genetic solutions. In every environment, DDPG+HER outperforms other algorithms (such as PPO, DDPG, and A2C). This might be because the addition of HER has improved DDPG's performance. Additionally, when rewards are sparse, DDPG+HER performs better \cite{plappert2018multi}. DDPG alone (without HER) cannot perform in all environments (except in \textit{FetchReach} and \textit{DoorOpening} environments) as it has zero success rate in these environments. We think this is because, for DDPG, dense rewards are often simpler to learn from, whereas sparse rewards are more challenging. The environment of \textit{FetchReach} is undoubtedly quite simple, and all configurations can successfully address it. DDPG is capable of performing in a \textit{DoorOpening} environment, although not as well as DDPG+HER and GA+DDPG+HER. With the exception of \textit{FetchReach}, all environments for PPO show a zero success rate. This might be the case because PPO functions better in discrete environments as opposed to continuous environments. This may also apply to A2C, which has a success rate of zero in all environments, with the exception of \textit{FetchReach} and \textit{AuboReach}. It can be concluded that GA+DDPG+HER outperforms all of these techniques. This demonstrates how performance may be improved by using GA for hyperparameter tuning. 

\section{Conclusion and Future Work}\label{sec5}
This study provided early data that showed how a genetic algorithm might enhance the hyperparameters of a reinforcement learning system to enhance performance, as demonstrated by quicker competence on six manipulation tasks. We reviewed previous work in reinforcement learning in robotics, presented the GA+DDPG+HER algorithm for reducing the number of epochs required to reach maximum performance, and described why a GA would be appropriate for such optimization. Our results on the six manipulation tasks show that the GA can identify hyperparameter values that lead to faster learning and higher (or equal) performance at our chosen tasks, confirming our hypothesis that GAs are a suitable fit for such hyperparameter optimization. We compared GA+DDPG+HER to other approaches and found ours as the most effective. In conclusion, the implementation supports the claim that real robots learn more quickly when hyperparameters for DDPG+HER are discovered using GA+DDPG+HER. Faster transitions between initial and target states were possible in the real robot environment, \textit{AuboReach}. Additionally, it may be claimed that GA+DDPG+HER acts similarly to its equivalent of DDPG+HER in terms of problem-solving speed but suffers from the same drawbacks when a problem may only have a limited number of genetic solutions, such as in the \textit{FetchSlide} environment.

We also showed that heuristic search, as implemented by genetic and other evolutionary computing techniques, is a feasible computational tool for improving reinforcement learning and sensor odometry performance. Adaptive genetic algorithms can also be implemented to take on different sets of hyperparameters during system execution. This could indicate online hyperparameter adjustment, which can help any system perform better, regardless of the domain or testing environment type.

\section*{Declarations}

\subsection*{Funding}
This work is supported by the U.S. National Science Foundation (NSF) under grants NSF-CAREER: 1846513 and NSF-PFI-TT: 1919127, and the U.S. Department of Transportation, Office of the Assistant Secretary for Research and Technology (USDOT/OST-R) under Grant No. 69A3551747126 through INSPIRE University Transportation Center, and the Vingroup Joint Stock Company and supported by Vingroup Innovation Foundation (VINIF) under project code VINIF.2020.NCUD.DA094. The views, opinions, findings, and conclusions reflected in this publication are solely those of the authors and do not represent the official policy or position of the NSF, USDOT/OST-R, and VINIF.

\subsection*{Conflicts of interest/Competing interests}
The authors declare that there is no conflict of interest.

\subsection*{Code/Data availability}
Open source code, and data used in the study is available at \textcolor{orange}{\href{https://github.com/aralab-unr/ga-drl-aubo-ara-lab}{https://github.com/aralab-unr/ga-drl-aubo-ara-lab}}.

\subsection*{Authors' contributions}
Adarsh Sehgal is the first author and primary contributor to this paper. Adarsh did the research and wrote this manuscript. Nicholas Ward assisted Adarsh in code execution and gathering results. Drs. Hung La and Sushil Louis advised and overlooked this study.

\subsection*{Ethics approval}
Not applicable

\subsection*{Consent to participate}
All authors are full-informed about the content of this paper, and the consent was given for participation.

\subsection*{Consent for publication}
All authors agree on the content of this paper for publication.


\bibliography{mybibfile}


\end{document}